\documentclass{article}

\usepackage{utils/arxiv}
\newcommand{\undertitle}{}
\usepackage[utf8]{inputenc} 
\usepackage[T1]{fontenc}    
\usepackage{hyperref}       
\usepackage{url}            
\usepackage{booktabs}       
\usepackage{amsfonts}       
\usepackage{nicefrac}       
\usepackage{microtype}      
\usepackage{amsmath}        
\usepackage{cleveref}       
\usepackage{lipsum}         
\usepackage{graphicx}
\usepackage{natbib}
\usepackage{doi}

\usepackage{graphicx}
\usepackage{caption}
\usepackage{subcaption} 
\usepackage{booktabs}
\usepackage{multirow}
\usepackage{tikz}
\usepackage{pgfplots}
\pgfplotsset{compat=1.18}
\usepackage[table]{xcolor}
\usepackage{colortbl}
\usepackage{subcaption}
\usepackage{array}
\usepackage{adjustbox}
\usepackage[utf8]{inputenc} %
\usepackage[T1]{fontenc}    %
\usepackage{url}            %
\usepackage{booktabs}       %
\usepackage{amsfonts}       %
\usepackage{nicefrac}       %
\usepackage{microtype}      %

\usepackage{bbding}
\usepackage{multirow}
\usepackage{algorithm}
\usepackage{graphicx, amsmath, amssymb, caption, subcaption, overpic, textpos}
\usepackage{arydshln} %
\usepackage{listings}
\usepackage{setspace}
\usepackage{xspace}
\usepackage{subcaption}
\usepackage{xspace}
\usepackage[table]{xcolor}
\usepackage{wrapfig}
\usepackage{colortbl}

\newcommand{\modelname}{FlowInOne\xspace}
\newcommand{\datasetname}{VisPrompt-5M\xspace}
\newcommand{\benchname}{VP-Bench\xspace}
\newcommand{\FM}{flow matching\xspace}  %
\usepackage{algorithm}
\usepackage{algpseudocode}
\usepackage{amsmath}

\definecolor{FlowBlue}{HTML}{0052CC}
\newcommand{\modelnamecolored}{%
    \textcolor{FlowBlue!100}{F}%
    \textcolor{FlowBlue!92}{l}%
    \textcolor{FlowBlue!85}{o}%
    \textcolor{FlowBlue!77}{w}%
    \textcolor{FlowBlue!70}{I}%
    \textcolor{FlowBlue!62}{n}%
    \textcolor{FlowBlue!55}{O}%
    \textcolor{FlowBlue!47}{n}%
    \textcolor{FlowBlue!40}{e}\xspace
}
\newcommand{\iio}{%
    \textcolor{FlowBlue!100}{I}%
    \textcolor{FlowBlue!96}{m}%
    \textcolor{FlowBlue!93}{a}%
    \textcolor{FlowBlue!89}{g}%
    \textcolor{FlowBlue!86}{e}%
    \textcolor{FlowBlue!82}{-}%
    \textcolor{FlowBlue!79}{i}%
    \textcolor{FlowBlue!75}{n}%
    \textcolor{FlowBlue!71}{,}\hspace{0.2em}%
    \textcolor{FlowBlue!68}{I}%
    \textcolor{FlowBlue!64}{m}%
    \textcolor{FlowBlue!61}{a}%
    \textcolor{FlowBlue!57}{g}%
    \textcolor{FlowBlue!54}{e}%
    \textcolor{FlowBlue!50}{-}%
    \textcolor{FlowBlue!47}{o}%
    \textcolor{FlowBlue!43}{u}%
    \textcolor{FlowBlue!40}{t}\xspace
}
\NewDocumentCommand{\inlineimage}{O{1.0} m}{%
  \raisebox{-0.15\baselineskip}{\includegraphics[height=#1\baselineskip]{#2}}\hspace{0.2em}
}
\newcommand{\flowicon}{\inlineimage[1.1]{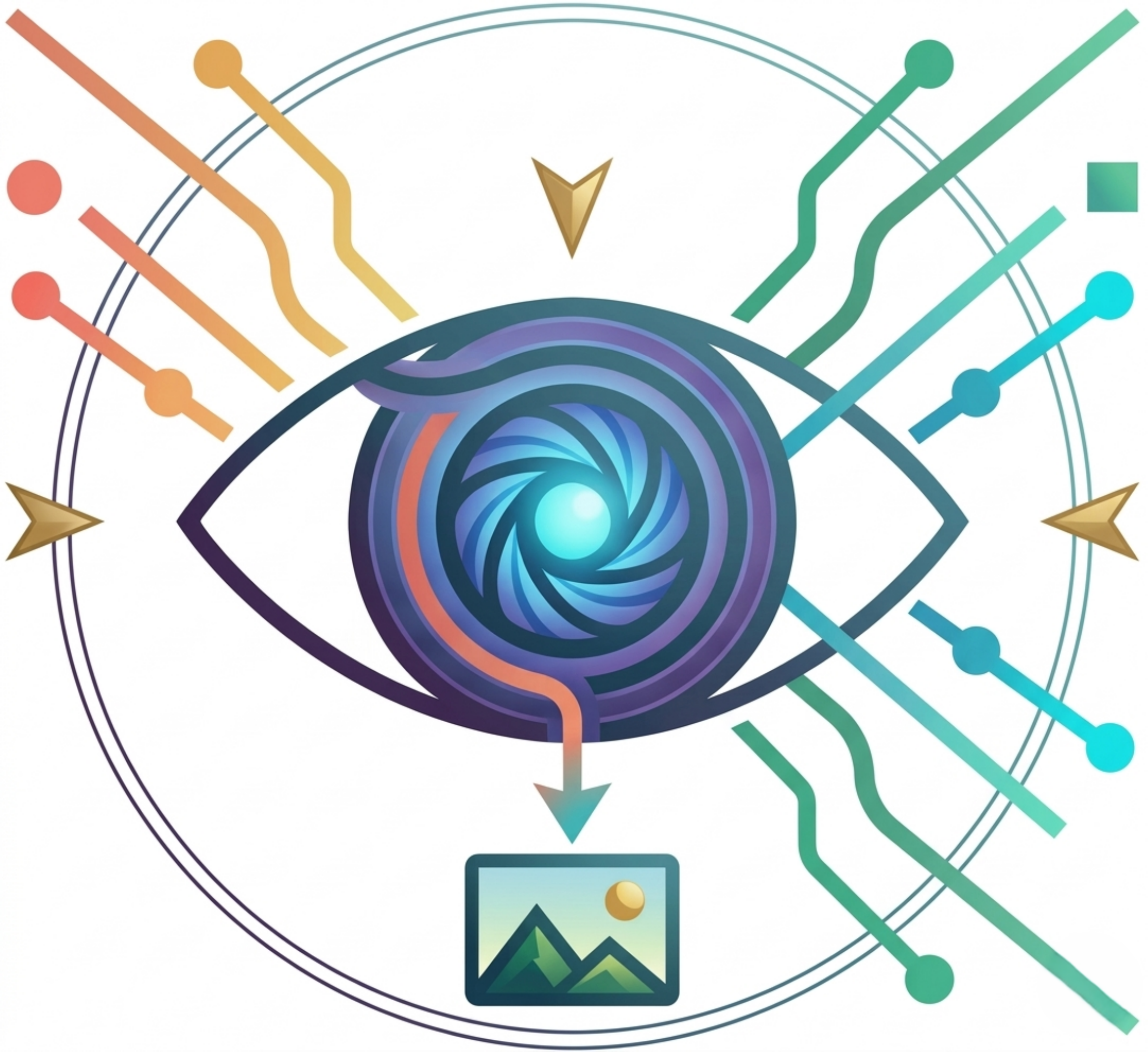}}
\newcommand{\best}[1]{\cellcolor{red!12}#1}
\title{\flowicon \modelnamecolored: Unifying Multimodal Generation as \\ \iio Flow Matching}

\date{}

\renewcommand{\shorttitle}{\textit{arXiv} Template}

\author{
Junchao Yi $^1$$^*$,~Rui Zhao $^3$$^*$,~Jiahao Tang $^2$$^*$,~Weixian Lei$^3$,~Linjie Li $^5$,
~Qisheng Su $^4$,
\\
\textbf{~Zhengyuan Yang $^5$,~Lijuan Wang $^5$, ~Xiaofeng Zhu $^1$$^\dagger$, Alex Jinpeng Wang $^2$$^\dagger$}
\\
$^1$ University of Electronic Science and Technology of China
$^2$ Central South University
\\
$^3$ National University of Singapore
$^4$ University of Science and Technology of China
$^5$ Microsoft
}

\begin{document}
\maketitle

\vspace{-35pt} 
\begin{center}
\texttt{Homepage: \url{https://csu-jpg.github.io/FlowInOne.github.io/}}
\end{center}
\vspace{10pt} 

\let\thefootnote\relax\footnotetext{$^*$Equal contribution, $^\dagger$Corresponding author}

\begin{abstract}
Multimodal generation has long been dominated by text-driven pipelines where language dictates vision but cannot reason or create within it. 
We challenge this paradigm by asking whether all modalities, including textual descriptions, spatial layouts, and editing instructions, can be unified into a single visual representation. 
We present \modelname, a framework that reformulates multimodal generation as a \textbf{purely visual flow}, converting all inputs into visual prompts and enabling a clean \textbf{image-in, image-out} pipeline governed by a single flow matching model. 
This vision-centric formulation naturally eliminates cross-modal alignment bottlenecks, noise scheduling, and task-specific architectural branches, unifying text-to-image generation, layout-guided editing, and visual instruction following under one coherent paradigm. 
To support this, we introduce \textbf{\datasetname}, a large-scale dataset of 5 million visual prompt pairs spanning diverse tasks including physics-aware force dynamics and trajectory prediction, alongside \textbf{\benchname}, a rigorously curated benchmark assessing instruction faithfulness, spatial precision, visual realism, and content consistency.
Extensive experiments demonstrate that \modelname achieves state-of-the-art performance among open-source models across all unified generation tasks while remaining competitive with leading commercial systems, thereby establishing a new foundation for fully vision-centric generative modeling, in which perception and creation coexist within a unified continuous visual space.
\end{abstract}
    
\section{Introduction}
\label{intro}
\begin{wrapfigure}{R}{0.5\textwidth}
  \centering
  \includegraphics[width=0.48\textwidth]{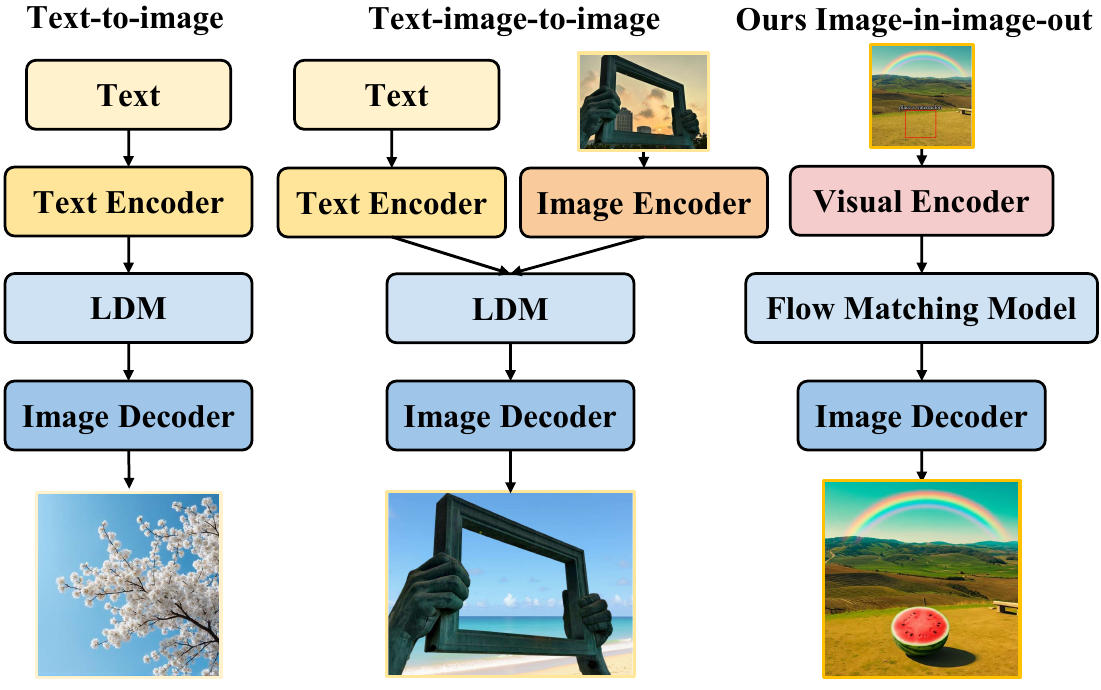} 
  \captionsetup{font=footnotesize}
  \caption{\textbf{Comparison of generation paradigms.} Left: Traditional T2I only uses the text encoder to condition the Latent Diffusion Model(LDM); Middle: Traditional TI2I requires the joint conditioning of both the text and image encoders; Right: We unify the conditions as visual input and form a simple image-in, image-out framework with a single model.}
  \label{fig:dif}
  \vspace{-4mm} 
\end{wrapfigure}
Multimodal generation has long operated under a text-dominant assumption: language encodes intent, and vision executes it. 
Models such as diffusion- and transformer-based text-to-image systems rely on linguistic embeddings as the central conditioning source.  
While this design achieves impressive visual fidelity, it introduces a fundamental asymmetry: \textit{language controls vision, but vision cannot reason or generate on its own}.  
This fragmentation of representation space makes it inherently difficult to unify understanding, editing, and generation within a single coherent model.

Recently, a growing trend of vision-centric models has emerged~\cite{robust_translation,PIXEL,xiao2024pixel}, suggesting that textual information can be processed in a purely visual manner by rendering text into pixel space.
These studies collectively suggest that the visual modality itself is expressive enough to serve as the foundation for multimodal understanding. 
Together, these studies show that representing language visually enables unified perception and alignment within a single modality.  
Yet they remain fundamentally \textit{perception-oriented}, leaving the generative potential of this vision-first formulation largely unexplored.
This raises an important question: \textbf{can we build a large model that both reasons and generates entirely within the visual space?}

Flow matching~\cite{lipman2023flow}  provides a principled answer to this question. 
Compared with diffusion, it directly learns the underlying velocity field of transformation, offering higher sampling efficiency and stable optimization.
By learning visual flows instead of stochastic noise removal, it connects perception and generation under a single deterministic principle.

In this work, we take a decisive step toward this goal and introduce \modelname, a framework that redefines multimodal generation as a \textit{purely visual flow}.
In \modelname, text, layout, and instruction inputs are first transformed into visual prompts, forming the input image state.
The model then learns a continuous transport process that evolves this state into the target visual output using flow matching.
This formulation enables a simple and general training pipeline that eliminates noise scheduling, diffusion sampling, and task-specific condition heads.

As shown in Figure~\ref{fig:dif}, \modelname departs from the traditional text-conditioned pipeline.
Conventional text-to-image or image editing models use text encoders (e.g., Flan-T5~\cite{flant5}) to condition a latent diffusion model, while text-image-to-image setups require two encoders for joint conditioning.
In contrast, \modelname unifies all input conditions as visual prompts, forming a simple image-in, image-out pipeline with a single model.
This design not only simplifies architecture but also ensures consistent alignment between semantic content and spatial control across diverse tasks such as text-to-image generation, layout-guided editing, and visual instruction following.

To support this unified paradigm, we construct \datasetname, a large-scale Visual Prompt Dataset that spans text-in-image generation, versatile visual editing and physics-aware instruction following.
Each sample pairs a visual prompt canvas with its corresponding target image, providing supervision as continuous visual evolution without task-specific modules or auxiliary channels. 
We further introduce \benchname, a carefully curated evaluation benchmark that assesses model performance across four dimensions: instruction faithfulness, content consistency, visual realism, and spatial precision.

Our main contributions are as follows. 
\emph{i}. We reformulate multimodal generation into a \textbf{vision-centric image-in, image-out paradigm}, eliminating the text encoders and modality-specific bridges.
\emph{ii}. We propose \modelname,  a unified flow matching framework that models multimodal transformation as continuous visual evolution within a shared latent space. 
\emph{iii}. We build \datasetname, a comprehensive dataset of visual prompts that enables unified training and strong generalization across text-to-image, image-to-image, and instruction-guided generation tasks.
Extensive experiments demonstrate that \modelname achieves state-of-the-art performance across unified generation, precise image editing, and physics-aware instruction following, establishing a new foundation for fully vision-centric generative modeling.
\section{Related Works}
\paragraph{Diffusion and Flow Matching.}
While diffusion models, from DDPM~\cite{ddpm,song2021ddim} to LDM~\cite{LDM_2022_CVPR} and DiT~\cite{Peebles2022ScalableDM}, dominate image generation via progressive denoising, Flow Matching~\cite{lipman2023flow,liu2023flow,geng2025meanflowsonestepgenerative,zhang2025alphaflowunderstandingimprovingmeanflow} learns a continuous transport map between distributions. 
This approach reduces reliance on complex noise schedules while enhancing sampling efficiency and stability~\cite{DiscreteFlowMatching}.
Building on this, \modelname directly models the continuous evolution within a shared latent space, entirely eliminating additional conditioning or noise injection.

\paragraph{Text- and Image-Conditioned Generation.}
Current T2I models~\cite{pmlr-v139-ramesh21a,Zhou2022ShiftedDF,podell2023sdxlimprovinglatentdiffusion,chen2024pixartalpha,sun2024autoregressive} typically inject discrete text tokens via cross-attention~\cite{esser2024scaling,polyak2025moviegencastmedia}, leaving control signals disjointed from the visual space.
Similarly, conventional image-to-image translation~\cite{xiao2025thermalgen,Nobis2025FractionalDB,Liu2023I2SBIS,liu2023flow,zhou2024denoising}, restoration~\cite{wang2025residualdiffusionbridgemodel,liu2025latentharmonysynergisticunified}, and controllable editing~\cite{Brooks_2023_CVPR,hertz2023prompttoprompt,li2023gligen,Zhang2023AddingCC,chen2023textdiffuser,yang2024glyphcontrol,pan2023draggan} rely heavily on adversarial frameworks~\cite{mirza2014conditionalgenerativeadversarialnets,zhu2020unpairedimagetoimagetranslationusing}, diffusion priors~\cite{meng2022sdedit}, or external control channels.
A common limitation across these methods is the dependence on explicit masks or task-specific interfaces for geometric and semantic control.
In contrast, \modelname entirely bypasses specialized conditioning branches. By rendering heterogeneous constraints—such as text and arrows, we converge multiple forms of control into a single image input, natively aligning semantics and geometry within the visual domain.
\paragraph{Modal mapping.}
While standard diffusion models map discrete text to images across divergent modalities~\cite{chen2024pixartalpha,fan2024fluid,ren2025xar,yu2024randomized,weber2024maskbit,kim2025democratizing,bai2025meissonic,chen2025hawkleveragingspatialcontext,zheng2025dense2moerestructuringdiffusiontransformer,chang2025maskattnsdxlcontrollableregionleveltexttoimage}, inherent modality gaps, tokenization artifacts, and reliance on Gaussian noise severely limit spatial precision~\cite{jia2025principlesapplicationscomprehensivesurvey}.
Moving beyond generic image-to-image translation~\cite{Liu2023I2SBIS,liu2023flow,zhou2024denoising}, we frame generation fundamentally as an \textbf{intra-modal transport problem}. 
By encoding both the visually-instructed input and the target image into a shared, isomorphic latent space, we learn a direct, noise-free flow between them. 
This pure single-modality mapping resolves structural mismatches at the latent level, demonstrating exceptional scalability across diverse editing and generation tasks.
\section{Dataset and Benchmark}
In this section, we detail the construction of the \textbf{\datasetname} dataset (Sec.~\ref{subsec:dataset}) and our evaluation benchmark, \textbf{\benchname} (Sec.~\ref{sec:benchandeva}).
\begin{figure*}[t]
    \centering
    \includegraphics[width=\textwidth]{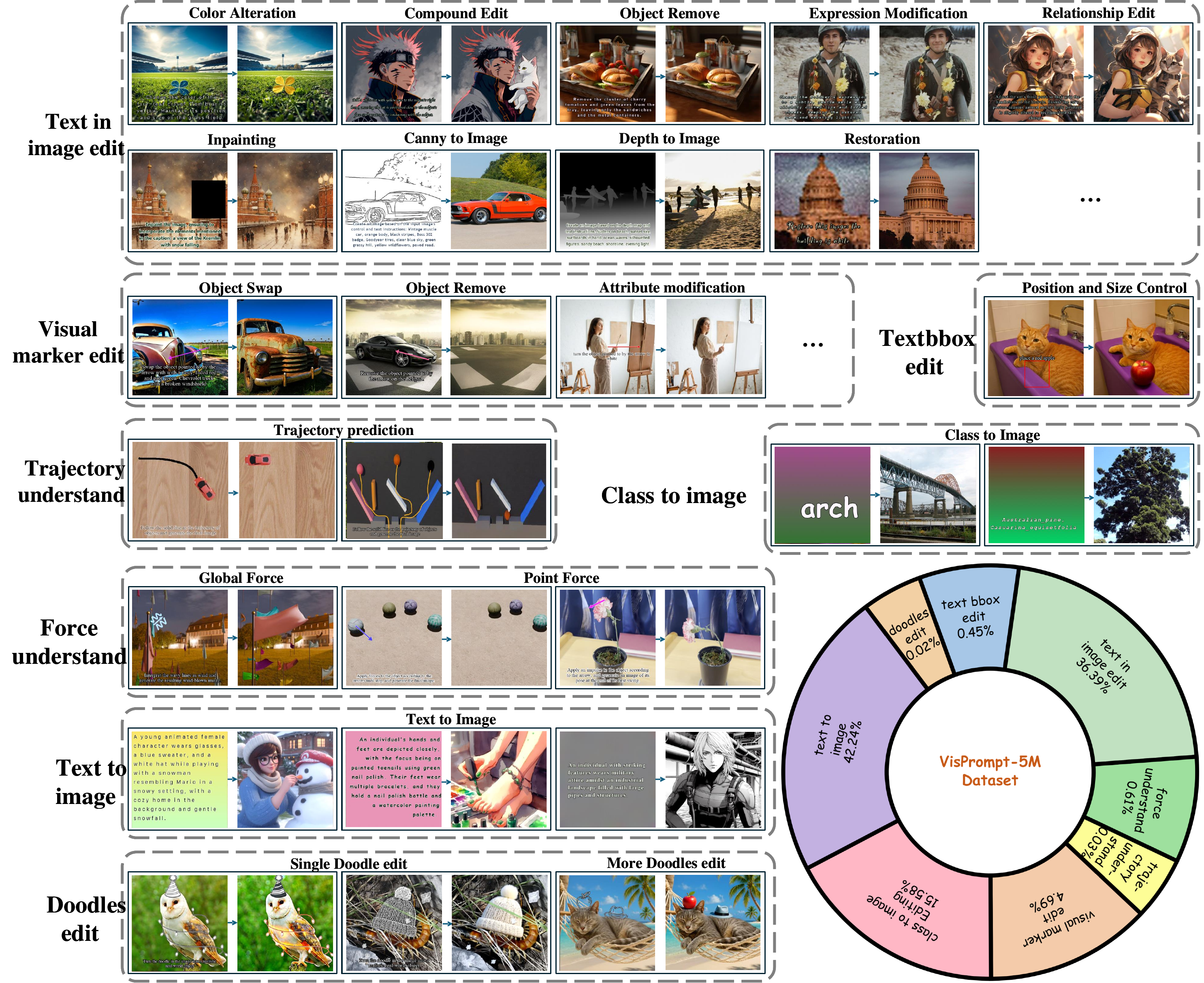}
    \caption{\datasetname is a comprehensive dataset that comprises eight distinct data types, including class-to-image generation, text-to-image generation, text-in-image editing, text bounding box editing, visual marker editing, doodles editing, force understanding, and trajectory understanding. The dataset covers a wide spectrum of image-to-image generation, ranging from basic text-in-image generation to compositional editing, and further to physics-aware instruction following.}
    \label{fig:dataset}
\end{figure*}
\subsection{\datasetname}
\label{subsec:dataset}
Illustrated in Figure~\ref{fig:dataset}, \datasetname enables a unified image-in, image-out paradigm. Training pairs $(I_v, I^\star)$ embed all textual and spatial instructions directly into the input canvas $I_v$. 
Eliminating auxiliary text channels mitigates ambiguity and enforces strict geometric alignment, empowering a single model to handle diverse tasks within one modality.

\noindent\textbf{Fundamental Generation.} We render textual prompts from text-to-image-2M~\cite{zk_2024} and class labels from an 860K high-quality ImageNet~\cite{Russakovsky2014ImageNetLS} subset directly as input canvases to match their corresponding target images.

\noindent\textbf{Text-in-Image Editing.} This category unifies diverse editing and condition-to-image tasks by overlaying textual instructions directly onto the input image canvas.
Drawing from GPT-Image-Edit~\cite{wang2025gptimageedit15mmillionscalegptgeneratedimage}, Pico-Banana~\cite{qian2025picobanana400klargescaledatasettextguided}, and UnicEdit~\cite{ye2025unicedit}, we filter out complex or inconsistent pairs to retain approximately 1.6M examples, spanning a diverse range of edits including additions, deletions, and attribute/environment changes.
Furthermore, we seamlessly integrate 315K structured image pairs from PixWizard~\cite{lin2024pixwizard}, which encompasses a broad spectrum of tasks including Canny-to-image, depth-to-image, inpainting, and image restoration. 
In total, this yields nearly 1.9M curated pairs where all operational intents are explicitly embedded as visual text prompts.

\noindent\textbf{Text Bounding Box Editing.} For precise object insertion with explicit geometric constraints, we extract a high-quality subset of 45K examples from GPT-Image-Edit. Leveraging the combined priors of Qwen Image Edit~\cite{wu2025qwenimagetechnicalreport} and Qwen3-VL~\cite{Qwen2.5-VL} to guide I2I generation, we synthesize pairs where text and bounding boxes jointly dictate the target category, scale, and location. Automated filtering via Qwen3-VL 7B ultimately retains 24K high-quality pairs.

\noindent\textbf{Visual Marker Editing.} Leveraging visual understanding capabilities of Qwen3-VL, we generate 250K pairs where arrow annotations serve as salient cues without requiring explicit object names in instruction. This subset supports  operations such as deletion, replacement, implicitly representing semantics and spatial relationships through visual markers.

\noindent\textbf{Doodles Editing.} Using a two-stage synthesis pipeline with Qwen Image Edit~\cite{wu2025qwenimagetechnicalreport} on 5K images crawled from the web, we first add doodles to create input images, then transform them into photo-realistic objects to form target images. Rigorous manual inspection mitigates generative instability, yielding 1K high-quality pairs where doodle lines explicitly serve as shape priors.

\noindent\textbf{Force \& Trajectory Understanding.} \datasetname supports physics-aware generation. For motion trajectories, we manually annotated the Blender-rendered videos of car and ball movements, yielding 1.5K strictly curated image pairs. For force understanding, we leverage the Force Prompting dataset~\cite{gillman2025forcepromptingvideogeneration} that covers aerodynamics, oscillations, and linear motion. By extracting keyframes and superimposing text and arrows to denote precise force magnitude and direction, we explicitly visualize object dynamics within the input image.

Across the entire pipeline, we enforce standardized data formats while strictly maintaining task metadata and geometric attributes. 
We also incorporate automated quality control measures via MLLMs to ensure semantic consistency, visual fidelity, and visual text readability.
\subsection{Benchmark and Evaluation}
\label{sec:benchandeva}
To evaluate our pure Image-in, Image-out paradigm, we curate \textbf{\benchname}, a comprehensive and manually filtered benchmark covering diverse visual instructions (details provided in Appendix B).

Since conventional metrics like FID~\cite{heusel2018ganstrainedtimescaleupdate} struggle with complex visual instructions, we follow recent studies~\cite{labs2025flux1kontextflowmatching, liu2025step1xeditpracticalframeworkgeneral} by adopting VLMs as our primary evaluators.
A generation is deemed successful only if it simultaneously satisfies four criteria: \textbf{(1) Instruction Faithfulness}, \textbf{(2) Content Consistency}, \textbf{(3) Visual Realism}, and \textbf{(4) Spatial Precision}.
Alongside the VLM assessment, we conduct rigorous human evaluation based on these exact same criteria to compute the overall pass rate.
Since standard VLMs may miss implicit constraints rendered on the image canvas, we manually extract the textual instructions and supply them as supplementary text prompts to ensure fair assessment (refer to Appendix F for VLM evaluation prompts).

To comprehensively evaluate visual quality and editing accuracy, we additionally tailor four quantitative metrics to our paradigm:
(1) \textbf{CLIP-IQA}~\cite{iqa} to measure overall visual realism;
(2) \textbf{CLIP Score}~\cite{clipscore} to evaluate semantic alignment, for which we directly extract the rendered text instructions from the input image and compute their similarity with the generated image;
(3) \textbf{Directional CLIP Similarity}~\cite{dir_clip} to assess semantic consistency in marker-based editing. Captions are manually generated for the input and generated images, and subsequently computing the directional alignment between the image transition and the corresponding caption pairs; and
(4) \textbf{DINOv3 Directional Similarity} ($\text{DINOv3 Sim}$) to accurately capture fine-grained spatial and physical structural changes, by directly computing the cosine similarity of the edit displacement vectors among the input, generated, and ground-truth images within the dense DINOv3~\cite{dinov3} feature space.
\section{Method}
In this section, we first briefly review the preliminaries of Flow Matching (Sec.~\ref{subsec:preliminaries_fm}) and introduce our core strategy for encoding visual instructions into a unified visual semantic space (Sec.~\ref{subsec:vis_seman}). 
Finally, we present the detailed architecture of \modelname (Sec.~\ref{subsec:model}), featuring a novel Dual-Path Spatially-Adaptive Modulation mechanism to balance structural preservation and instruction adherence.
\subsection{Preliminaries: Flow Matching}
\label{subsec:preliminaries_fm}
Flow Matching (FM)~\cite{lipman2023flow,liu2023flow,liu2024instaflowstephighqualitydiffusionbased} formulates generative modeling as a continuous transport from a source distribution $p_0$ to a target distribution $p_1$ over $t \in [0,1]$.
Unlike traditional diffusion models~\cite{ddpm,song2021ddim}, FM does not rely on complex noise scheduling and permits non-Gaussian source distributions, provided they are isomorphic to the target. 
During training, FM constructs a differentiable probability path $z_t$ between a sample pair $(z_0, z_1)$, which directly yields the ground-truth velocity $v_t^\star$ for supervision:
\begin{equation} \label{eq:fm}
    z_t = t z_1 + \left(1-(1-\sigma_{\min})t\right)z_0, \quad v_t^\star = \frac{\partial z_t}{\partial t} = z_1 - (1-\sigma_{\min})z_0
\end{equation}
The network learns a time-dependent velocity field $v_\theta(z_t,t)$ by minimizing the Mean Squared Error (MSE) against $v_t^\star$. 
In our \modelname framework, we explicitly define a non-Gaussian source distribution: $z_0$ represents the latent state of the unified visual instruction extracted via a visual encoder and text-image VAE, while $z_1$ represents the target image latent. 
Since both latents are isomorphic within a shared space, inference is intuitively performed by solving the Ordinary Differential Equation (ODE) $\frac{d z_t}{d t} = v_\theta(z_t, t)$ from $t=0$ to $t=1$, deterministically evolving the visual instruction into the final target image.
\subsection{Unified processing of text in visual semantic space}
\label{subsec:vis_seman}
The inherent heterogeneity between discrete linguistic symbols and continuous visual textures poses significant alignment challenges in flow matching.
To alleviate this, we propose a paradigm shift: rendering textual instructions and diverse visual cues directly onto the image canvas. 
This explicitly preserves spatial layouts and structural priors without relying on complex cross-modal alignment modules.

By treating text as an integral part of the visual geometry, we circumvent the semantic fragmentation typically introduced by textual tokenizers.
To extract robust representations from this unified image $I_v$, we leverage the visual encoder of Janus-Pro-1B~\cite{chen2025januspro}.
The input is processed by a SigLIP Vision Transformer to extract patch-level semantic features, which are then mapped into the target embedding space via an MLP projector:
\begin{equation}
X_{\text{fuse}} = \text{MLP}(\text{SigLIP}(I_{v})) \in \mathbb{R}^{N \times D}
\end{equation}
where $N$ is the number of patches and $D$ denotes the embedding dimension. 
This sequence, $X_{\text{fuse}}$, encapsulates both textual semantics and visual geometry.

To perform continuous flow matching, we map the unified visual tokens to a source latent space via a text-image VAE. 
Rather than predicting it directly, it parameterizes a distribution to sample the source state $Z_{TI} \sim \mathcal{N}(\bar{\mu}_{z_{0}}, \operatorname{diag}(\bar{\sigma}_{z_{0}}^{2})) \in \mathbb{R}^{H \times W \times C}$. 
Symmetrically, a frozen image VAE encodes the target image into an isomorphic latent $Z_{I}$.
Our generative process is thus elegantly formulated as modeling the time-dependent velocity field that continuously transports the source latent $Z_{TI}$ to the target image latent $Z_{I}$ within the shared latent space.

\subsection{\modelname}
\label{subsec:model}
\begin{figure*}
    \centering
    \includegraphics[width=\linewidth]{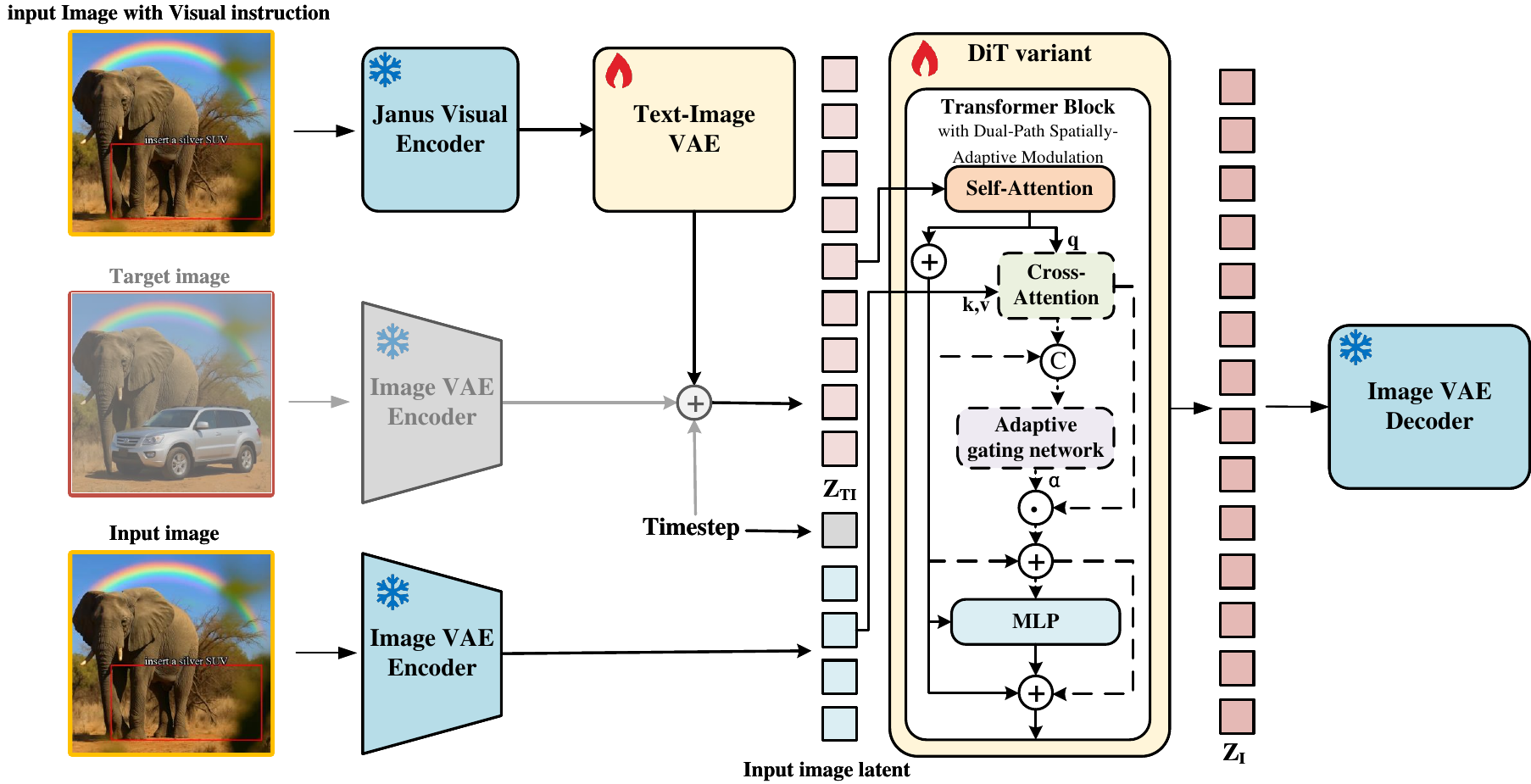}
    \caption{\textbf{Overview of the \modelname architecture, a general and simple framework using \FM for continuous evolution in only one modality.}
    \modelname employs a Dual-Path Spatially-Adaptive Modulation to adapt computation by modality. For input image rendering with only text, the structural branch is bypassed to strictly follow semantic evolution. Conversely, for image editing, a spatially-adaptive gated network and cross attention activates to selectively inject source priors, dynamically balancing original image preservation with instruction-driven reconstruction.
    }
    \label{fig:arch}
\end{figure*}
\paragraph{Dual-Path Spatially-Adaptive Modulation.}
Within the unified Flow Matching framework proposed in this work, we formulate image generation as a deterministic trajectory evolution from a starting distribution $p_0$ to a target distribution $p_1$. 
However, due to the information compression inherent in the visual encoding stage, the initial latent ${Z}_{TI}$ often fails to fully capture the fine-grained structural features of the source image ${I}_{src}$. 
To address this, we introduce a Dual-Path Spatially-Adaptive Modulation mechanism. 
This mechanism is designed to dynamically compensate for the missing structural manifold while switching computational paths based on the specific task type.
As shown in Figure~\ref{fig:arch}, let $\mathbf{H}^{(l)} \in \mathbb{R}^{N \times D}$ denote the hidden state of the $l$-th Transformer layer, where $N$ is the token sequence length and $D$ is the feature dimension. 
This state is first updated via a self-attention $\mathcal{A}_{self}$ to capture the global contextual information:
\begin{equation}
\begin{aligned}
\tilde{\mathbf{H}}^{(l)} = \text{LayerNorm}(\mathbf{H}^{(l)} + \mathcal{A}_{self}(\mathbf{H}^{(l)}))
\end{aligned}
\end{equation}
Upon obtaining the features enhanced by self-attention, the model follows a dual-path conditional branch defined by the two different tasks. 
For text-to-image generation in visual input, where no external structural prior needs to be maintained, the model bypasses the cross-attention layer to prevent the introduction of irrelevant noise. 
This ensures that the generation trajectory strictly follows the evolution dictated by the text semantics. 
Conversely, for image editing involving a source image, we employ an Image VAE Encoder to map ${I}_{src}$ into the latent space, resulting in the latent $\mathbf{z}_{src} \in \mathbb{R}^{H \times W \times C}$. 
Since $\mathbf{z}_{src}$ captures the visual structure of the image, we reshape it into a reference sequence $\mathbf{S} \in \mathbb{R}^{N \times C}$, where $N = H \times W$. 
Subsequently, the structural increment $\Delta \mathbf{H}_{struct}$ is computed via a cross-attention mechanism, where the intermediate state $\tilde{\mathbf{H}}^{(l)}$ serves as the Query and the reference sequence $\mathbf{S}$ acts as the Key-Value pair:
\begin{equation}
\begin{aligned}
\Delta \mathbf{H}_{struct} = \text{Softmax}\left(\frac{(\tilde{\mathbf{H}}^{(l)} \mathbf{W}_Q)(\mathbf{S} \mathbf{W}_K)^\top}{\sqrt{d_k}}\right)(\mathbf{S} \mathbf{W}_V)
\end{aligned}
\end{equation}
To achieve an optimal trade-off between source image fidelity and instruction-following alignment, we design a lightweight adaptive gating network. 
By concatenating the current denoising state with the extracted source information, the network predicts an anisotropic token-level weight vector $\boldsymbol{\Lambda} \in [0, 1]^N$. 
This design enables the model to identify spatial heterogeneity at a pixel-level granularity—specifically, distinguishing between regions belonging to the background manifold that require strict preservation and regions targeted for editing that require reconstruction. 
The derivation of the gating coefficient matrix is as follows:
\begin{equation}
\begin{aligned}
\boldsymbol{\Lambda} = \sigma \left( \text{MLP}_{\theta} \left( [ \tilde{\mathbf{H}}^{(l)} \parallel \Delta \mathbf{H}_{struct} ] \right) \right)
\end{aligned}
\end{equation}
where $\parallel$ denotes feature concatenation along the channel axis and $\sigma$ represents the Sigmoid activation function. 
The final layer output $\mathbf{H}_{out}^{(l)}$ is integrated via a conditional formulation controlled by a task indicator $\mathbb{I}_{edit}$:
\begin{equation}
\begin{aligned}
\mathbf{H}_{out}^{(l)} = \tilde{\mathbf{H}}^{(l)} + \mathbb{I}_{edit} \cdot \left( \boldsymbol{\Lambda} \odot \Delta \mathbf{H}_{struct} \right)
\end{aligned}
\end{equation}
Here, $\mathbb{I}_{edit} \in \{0, 1\}$ is a binary indicator: for pure text-to-image inputs, $\mathbb{I}_{edit}=0$, and the modulation term is nullified to sever structural dependencies; for inputs containing visual instruction and source images, $\mathbb{I}_{edit}=1$, activating the spatially-adaptive refinement.
By precisely controlling the infiltration of the structural manifold via $\boldsymbol{\Lambda}$, this method effectively mitigates editing conflicts caused by over-preservation. 
Furthermore, by leveraging explicit structural compensation, it significantly reduces the evolution error of Flow Matching in complex image editing scenarios.

\paragraph{Flowing in the unified modality.}
We model visual instruction images  to target images as a flow matching process in a shared latent space. 
Given the instruction sequence \(X \in R^{N\times C}\) obtained from a unified visual encoder, a variational posterior is defined using a 
Variational Autoencoder. 
\begin{equation}
\begin{aligned}
q_\psi(z\,|\,X) &= \mathcal{N}\!\left(\bar{\mu}_X,\; \operatorname{diag}\!\left(\bar{\sigma}_X^{2}\right)\right),\\
z_0 &\sim q_\psi(z\,|\,X),\quad z_0\in\mathbb{R}^{N\times D}.
\end{aligned}
\end{equation}
Image-side training and sampling are conducted in a latent space: the pre-trained and frozen VAE from LDM~\cite{LDM_2022_CVPR} \(\mathrm{Enc}_{\mathrm{img}}\) is used to map the target image \(I^\star\) to the target latent variable.
\begin{equation}
\begin{aligned}
z_1=\mathrm{Enc}_{\mathrm{img}}(I^\star)\in\mathbb{R}^{N\times D}.
\end{aligned}
\end{equation}
During training, vanilla flow matching is adopted: time is sampled from \(t \sim \mathcal{U}(0, 1)\) and linear interpolation is constructed in Equation~\ref{eq:fm}.
The instantaneous velocity is represented by the vector field \(v_\theta(z,t)\), minimizing 
\begin{equation}
\label{equ:Lfm}
\begin{aligned}
\mathcal{L}_{\mathrm{FM}}=\mathbb{E}_{(z_0,z_1),t}\big|\big| v_\theta(z_t,t)-(z_1-z_0)\big|\big|_2^{2}.
\end{aligned}
\end{equation}
During inference, given only the visual instruction image \(I_v\), first take \(z_0\sim q_\psi(\cdot|E_v(I_v))\), and then solve the ordinary differential equation. 
Obtaining \(\hat{z}_1=z_{t=1}\), the final image \(\hat{I}=\mathrm{Dec}_{\mathrm{img}}(\hat{z}_1)\) is generated through the frozen image VAE decoder. 
This process achieves continuous transportation from the instruction state to the image state within a unified one-dimensional sequence modality, avoiding additional noise scheduling and conditional branching, and maintaining an isomorphic representation to the visual instructions.
\section{Experiment Results}
In this section, we first provide the implementation details of \modelname (Sec.~\ref{subsec:implementation_details}), and then present the main results on our carefully curated \benchname (Sec.~\ref{subsec:eval}). 
Finally, we conduct ablation studies to better understand the design choices of \modelname for image generation (Section~\ref{subsec:abla}).
\subsection{Implementation Details}
\label{subsec:implementation_details}
\paragraph{Model architecture.}
Based on the CrossFlow framework~\cite{liu2025flowing}, we encode unified image input via Janus-pro-1B~\cite{chen2025januspro} and a frozen LDM VAE~\cite{LDM_2022_CVPR}, projecting them through a stacked Transformer text-image VAE.
We augment the Transformer blocks with additional cross-attention layers.
The concatenated self- and cross-attention outputs are then fed into a lightweight network to predict spatially adaptive weights for input feature modulation.
\paragraph{Training details.}
The 1.2B \modelname is initialized from CrossFlow~\cite{liu2025flowing} and trained at $256 \times 256$ resolution for 240k steps via a balanced WebDataset~\cite{aizman2020highperformanceiolarge}, optimizing a combined Flow matching, KL divergence, and CLIP contrastive loss (see Appendix G for detailed configurations and more ablations).
\paragraph{Evaluation metrics.}
We evaluate \modelname on a high-quality subset of \benchname. Following Section~\ref{sec:benchandeva}, we report pass rates assessed by Gemini 3~\cite{gemini}, GPT 5.2~\cite{gpt}, and Qwen3.5~\cite{qwen3.5} and compute four quantitative metrics. Additionally, for qualitative verification, ten independent evaluators cross-checked 250 stratified random samples (25 each) to ensure judgment reliability.
\subsection{Evaluation on Image-to-image generation}
\label{subsec:eval}
\paragraph{State-of-the-art Comparison.}
We compare \modelname against several competitive open-source frameworks such as OmniGen2~\cite{wu2025omnigen2}, Qwen-Image-Edit-2509~\cite{wu2025qwenimagetechnicalreport}, and FLUX.1-Kontext-dev~\cite{labs2025flux1kontextflowmatching}, as well as the commercial model Nano Banana~\cite{nanobanana}. 
To ensure fairness,  we evaluate each baseline models through its optimal native interface, using detailed text prompts expanded by Qwen3-VL~\cite{Qwen2.5-VL} to supplement the image inputs.
Quantitative results demonstrate that \modelname possesses a significant advantage in unified generation tasks.
\begin{table}[!t]
  \centering
  \caption{
  Evaluation on the VP-Bench visual instruction benchmark using three VLM evaluators and human judges.  
  We use Gemini3~\cite{gemini}, GPT5.2~\cite{gpt} and Qwen3.5~\cite{qwen3.5} to evaluate the success ratio of \modelname on all kinds of visual instruction tasks. C2I: class to image, T2I: text to image, TIE: text in image edit, FU: force understanding, TBE: text \& bbox edit, TU: trajectory understanding, VME: visual marker edit, DE: doodles edit. \textbf{Total} denotes the average success rate across all sub-categories.}
  \label{tab:quan-vlm-vertical}
  \small 
  \tabcolsep=4pt
  \resizebox{0.9\textwidth}{!}{
  \begin{tabular}{l|cccccccc|c} 
    \toprule[0.15em]
    \textbf{Method} & \textbf{C2I} & \textbf{T2I} & \textbf{TIE} & \textbf{FU} & \textbf{TBE} & \textbf{TU} & \textbf{VME} & \textbf{DE} & \textbf{Total} \\
    \midrule
    \multicolumn{10}{c}{\textit{Evaluator: \textbf{Gemini3}}} \\
    \midrule
    Nano Banana~\cite{nanobanana} & .810 & \textbf{.980} & \textbf{.521} & .500 & \textbf{.600} & .020 & \textbf{.537} & \textbf{.740} & \textbf{.589} \\
    Omnigen2~\cite{wu2025omnigen2} & .720 & .760 & .313 & .013 & .020 & .000 & .020 & .140 & .248 \\
    Kontext~\cite{labs2025flux1kontextflowmatching} & .620 & .700 & .363 & .027 & .163 & .020 & .096 & .180 & .271 \\
    Qwen-IE-2509~\cite{wu2025qwenimagetechnicalreport} & .680 & .690 & .383 & .047 & .060 & .000 & .040 & .160 & .258 \\
    \textbf{FlowInOne (Ours)} & \textbf{.890} & .700 & .355 & \textbf{.727} & .302 & \textbf{.520} & .292 & .535 & .540 \\
    
    \midrule
    \multicolumn{10}{c}{\textit{Evaluator: \textbf{GPT5.2}}} \\
    \midrule
    Nano Banana~\cite{nanobanana} & .760 & \textbf{.960} & \textbf{.402} & .163 & .100 & .020 & \textbf{.227} & \textbf{.495} & .391 \\
    Omnigen2~\cite{wu2025omnigen2} & .660 & .820 & .203 & .001 & .000 & .000 & .001 & .160 & .231 \\
    Kontext~\cite{labs2025flux1kontextflowmatching} & .620 & .690 & .266 & .013 & .093 & .000 & .056 & .160 & .237 \\
    Qwen-IE-2509~\cite{wu2025qwenimagetechnicalreport} & .640 & .680 & .286 & .040 & .020 & .020 & .020 & .140 & .231 \\
    \textbf{FlowInOne (Ours)} & \textbf{.850} & .800 & .079 & \textbf{.500} & \textbf{.116} & \textbf{.240} & .083 & .465 & \textbf{.392} \\

    \midrule
    \multicolumn{10}{c}{\textit{Evaluator: \textbf{Qwen3.5}}} \\
    \midrule
    Nano Banana~\cite{nanobanana} & .780 & \textbf{.960} & \textbf{.446} & .427 & .260 & .040 & \textbf{.395} & \textbf{.760} & \textbf{.508} \\
    Omnigen2~\cite{wu2025omnigen2} & .740 & .790 & .257 & .027 & .020 & .000 & .010 & .160 & .251 \\
    Kontext~\cite{labs2025flux1kontextflowmatching} & .720 & .690 & .322 & .020 & .133 & .040 & .083 & .140 & .269 \\
    Qwen-IE-2509~\cite{wu2025qwenimagetechnicalreport} & .780 & .710 & .345 & .107 & .060 & .000 & .043 & .180 & .278 \\
    \textbf{FlowInOne (Ours)} & \textbf{.859} & .720 & .354 & \textbf{.713} & \textbf{.272} & \textbf{.320} & .306 & .481 & .503 \\
    
    \midrule
    \multicolumn{10}{c}{\textit{Evaluator: \textbf{Human}}} \\
    \midrule
    Nano Banana~\cite{nanobanana} & .790 & \textbf{.940} & \textbf{.372} & .287 & .220 & .020 & \textbf{.306} & \textbf{.740} & \textbf{.459} \\
    Omnigen2~\cite{wu2025omnigen2} & .720 & .710 & .268 & .013 & .020 & .000 & .010 & .120 & .233 \\
    Kontext~\cite{labs2025flux1kontextflowmatching} & .640 & .680 & .317 & .013 & .080 & .020 & .048 & .120 & .240 \\
    Qwen-IE-2509~\cite{wu2025qwenimagetechnicalreport} & .700 & .665 & .331 & .047 & .020 & .000 & .023 & .160 & .243 \\
    \textbf{FlowInOne (Ours)} & \textbf{.800} & .645 & .242 & \textbf{.705} & \textbf{.255} & \textbf{.280} & .255 & .400 & .449 \\
    \bottomrule[0.15em]
  \end{tabular}
  }
  \label{tab:quan-vlm}
\end{table}

As presented in Table~\ref{tab:quan-vlm}, \modelname consistently achieves the best performance among all open-source baselines across different evaluators. 
Specifically, \modelname obtains total success rates of \textbf{54.0\%}, \textbf{39.2\%}, \textbf{50.3\%}, and \textbf{44.9\%} under Gemini3, GPT5.2, Qwen3.5, and Human evaluation, respectively, substantially outperforming OmniGen2, FLUX.1-Kontext-dev, and Qwen-IE-2509. 
Notably, \modelname also remains highly competitive with the commercial model Nano Banana, achieving the highest total score under GPT5.2 and only slightly trailing Nano Banana under Gemini3, Qwen3.5, and Human evaluation. 
These results demonstrate that \modelname establishes a strong open-source baseline and approaches commercial-level performance in the image-in, image-out generation paradigm.

Table~\ref{tab:four_dimension_breakdown} further provides a fine-grained four-dimensional analysis. 
\modelname achieves consistently strong performance in instruction faithfulness and content consistency, showing that it can effectively follow visual instructions while preserving the required semantic content. 
More importantly, \modelname obtains the best spatial precision scores across all three evaluators, i.e., 3.42 under Gemini3, 3.24 under GPT5.2, and 3.30 under Qwen3.5, demonstrating its advantage in spatially grounded image-in, image-out generation. 
Although Nano Banana achieves higher visual realism, \modelname shows a better balance between instruction following and spatial control, especially compared with existing open-source baselines.
\begin{table}[!t]
\centering
\caption{
\textbf{Overall four-dimensional breakdown.}
Each dimension has a maximum score of 5 points.
IF: Instruction Faithfulness, CC: Content Consistency, VR: Visual Realism, SP: Spatial Precision.
}
\label{tab:four_dimension_breakdown}
\small
\setlength{\tabcolsep}{3pt}
\renewcommand{\arraystretch}{1.0}
\begin{tabular}{l|cccc|cccc|cccc}
  \toprule
  \multirow{2}{*}{\textbf{Method}}
  & \multicolumn{4}{c|}{\textbf{Gemini3}}
  & \multicolumn{4}{c|}{\textbf{GPT5.2}}
  & \multicolumn{4}{c}{\textbf{Qwen3.5}} \\
  \cmidrule(lr){2-5}
  \cmidrule(lr){6-9}
  \cmidrule(lr){10-13}
  & \textbf{IF} & \textbf{CC} & \textbf{VR} & \textbf{SP}
  & \textbf{IF} & \textbf{CC} & \textbf{VR} & \textbf{SP}
  & \textbf{IF} & \textbf{CC} & \textbf{VR} & \textbf{SP} \\
  \midrule
  Nano Banana   & \best{3.43} & \best{3.10} & \best{4.43} & 2.99 & \best{3.28} & 2.78 & \best{4.06} & 2.86 & 3.30 & \best{3.01} & \best{4.43} & 2.94 \\
  Omnigen2      & 2.15 & 1.39 & 3.46 & 1.69 & 2.29 & 1.53 & 3.31 & 1.92 & 1.72 & 0.91 & 3.13 & 1.27 \\
  Kontext       & 2.31 & 1.62 & 3.15 & 2.25 & 2.40 & 1.71 & 3.25 & 2.21 & 1.97 & 1.87 & 3.11 & 1.76 \\
  Qwen-IE-2509  & 2.96 & 2.51 & 3.28 & 1.91 & 2.76 & 2.39 & 3.58 & 1.94 & 2.33 & 1.98 & 3.23 & 1.39 \\
  \textbf{FlowInOne} 
                & 3.38 & 2.94 & 3.12 & \best{3.42} & 3.16 & \best{2.81} & 2.96 & \best{3.24} & \best{3.31} & 2.87 & 3.20 & \best{3.30} \\
  \bottomrule
\end{tabular}
\end{table}

Although minor discrepancies exist between MLLMs and human assessments due to limitations in grounding fine-grained visual markers~\cite{dong2025seeingreasoningmvpbenchgraphbased}, the overall ranking trends consistently align, verifying the reliability of our automated metrics.

Beyond pass rates, Table~\ref{tab:metric_results} further substantiates our findings. 
Crucially, \textbf{\modelname excels in fine-grained spatial and physical controls, achieving the highest average DINOv3 Sim score of 48.7\% (outperforming Nano Banana's 47.3\%)}, with notable margins in force \& trajectory understanding and text bbox editing. 
Furthermore, \textbf{in overall visual realism and semantic alignment, \modelname significantly surpasses all open-source baselines and performs comparably to the commercial model}, underscoring its robust generation quality and precise instruction-following capabilities.
\begin{table}[t]
\centering
\caption{Quantitative evaluation on VP-Bench across five models. CLIP-IQA is averaged over all categories to measure overall visual realism. CLIP Score evaluates semantic alignment for pure generation tasks. 
Directional CLIP Similarity assesses instruction-driven consistency in marker-based editing. DINOv3 Similarity measures fine-grained spatial and physical structural accuracy. 
}
\label{tab:metric_results}
\resizebox{\textwidth}{!}{%
\begin{tabular}{l |c |ccc |ccc |ccccc}
\toprule
\multirow{2}{*}{\textbf{Method}} & \textbf{IQA} $\uparrow$ & \multicolumn{3}{c|}{\textbf{CLIP Score} $\uparrow$} & \multicolumn{3}{c|}{\textbf{Dir CLIP} $\uparrow$} & \multicolumn{5}{c}{\textbf{DINOv3 $\text{Sim}$} $\uparrow$} \\
\cmidrule(lr){2-2} \cmidrule(lr){3-5} \cmidrule(lr){6-8} \cmidrule(lr){9-13}
 & Total & C2I & T2I & \bf Avg. & TIE & VME & \bf Avg. & DE & FU & TBE & TU & \bf Avg. \\
\midrule
Nano Banana~\cite{nanobanana} & \bf0.688 & 0.281 & \bf0.302 & \bf0.291 & \bf0.106 & \bf0.103 & \bf0.105 & \bf0.430 & 0.474 & 0.501 & 0.486 & 0.473 \\
Omnigen2~\cite{wu2025omnigen2} & 0.603 & 0.173 & 0.208 & 0.191 & 0.001 & 0.005 & 0.003 & 0.224 & 0.066 & 0.004 & 0.207 & 0.125 \\
Kontext~\cite{labs2025flux1kontextflowmatching} & 0.621 & 0.164 & 0.172 & 0.168 & 0.010 & 0.008 & 0.009 & 0.283 & 0.240 & 0.118 & 0.004 & 0.161 \\
Qwen-IE-2509~\cite{wu2025qwenimagetechnicalreport} & 0.646 & 0.231 & 0.216 & 0.224 & 0.005 & 0.011 & 0.008 & 0.133 & 0.215 & 0.250 & 0.207 & 0.201 \\
\textbf{FlowInOne (Ours)} & 0.684 & \bf0.290 & 0.276 & 0.283 & 0.092 & 0.101 & 0.097 & 0.335 & \bf0.536 & \bf0.506 & \bf0.570 & \bf0.487\\
\bottomrule
\end{tabular}
}
\end{table}

\begin{figure*}[t]
  \centering
  \includegraphics[width=\linewidth]{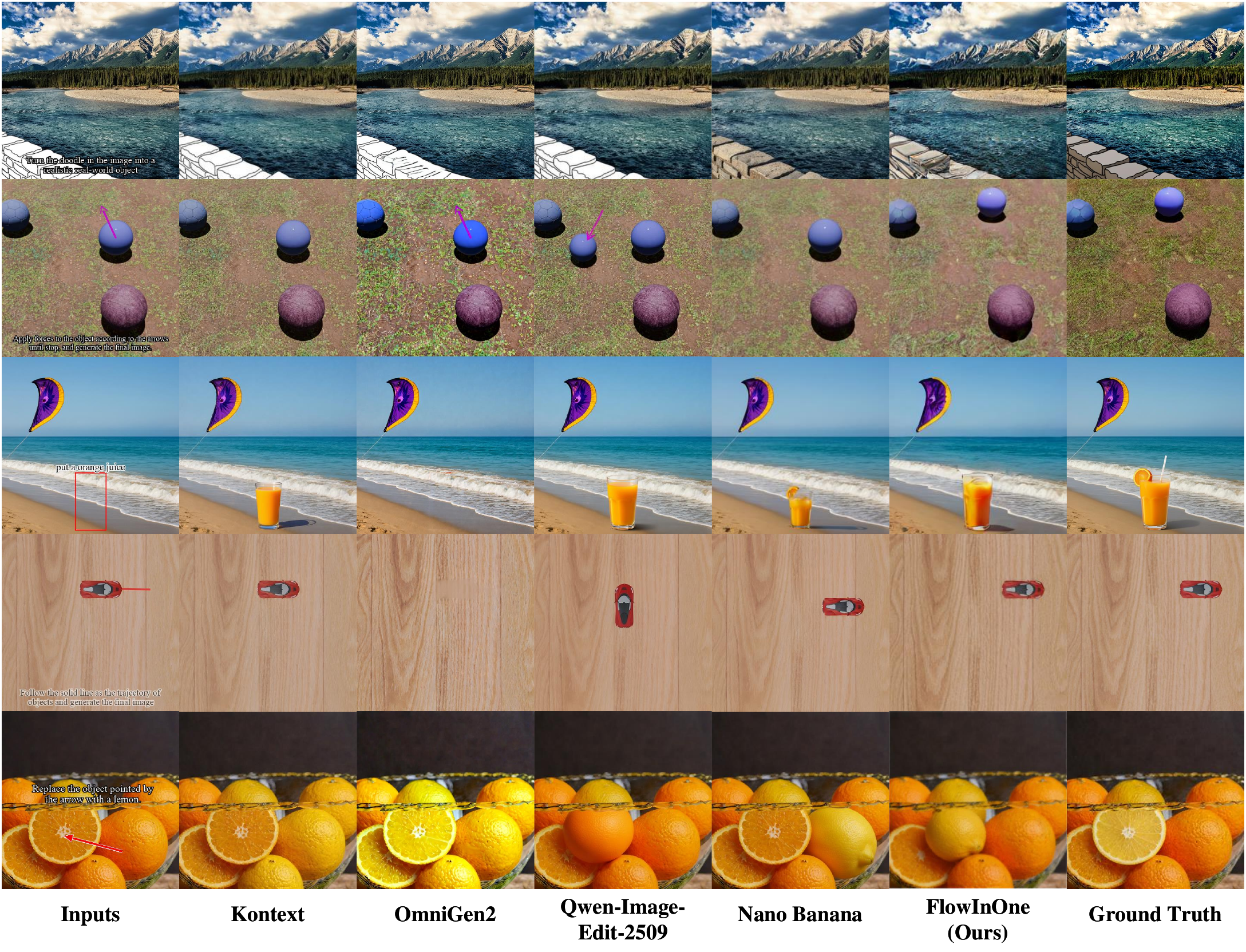}
  \caption{
    \textbf{Visual instruction editing comparison across methods.}
  }
  
  \label{fig:vis_comp}
  \vspace{-5mm}
\end{figure*}

\paragraph{Qualitative Comparison.}
Figure~\ref{fig:vis_comp} visually compares \modelname and baselines across five VP-Bench tasks: force, trajectory, text \& bbox, visual marker, and doodle editing. 
Unlike traditional pipelines, \modelname directly processes a unified canvas containing textual instructions, spatial layouts, and visual cues (arrows, markers, doodles). 
For fairness, we evaluate baselines via their optimal native interfaces: instructions are extracted and expanded into detailed prompts by Qwen3-VL before being provided alongside the processed image.

Despite these enhanced prompts, baselines often fail to translate visual cues into precise edits.
In force and trajectory tasks, they struggle to convert arrows into physically plausible motions. For text \& bbox editing, they frequently violate spatial constraints or size specifications. 
In marker and doodle tasks, baselines often misinterpret localized cues as scene elements, resulting in inaccurate synthesis or artifact retention.
Conversely, \modelname accurately executes modifications while preserving background consistency. 
This success confirms that our image-in, image-out paradigm enables more reliable grounding of fine-grained spatial and physical intents than conventional text-driven interfaces.

\subsection{Ablation Studies}
\label{subsec:abla}
We conduct a series of ablation studies to validate the architectural designs of our model.
Due to computational constraints, all models are trained for 100k steps with a batch size of 512 unless otherwise specified. 
To accurately reflect model performance, the reported Pass Rate averages the Gemini and GPT evaluations.
Furthermore, we adopt a progressive strategy, building upon the optimal configuration from each preceding part.
\begin{table}[htbp]
  \centering
  \caption{\textbf{Ablation study} on compression method, modulation method, and training strategy. The average scores across Gemini3~\cite{gemini}, GPT5.2~\cite{gpt} and Qwen3.5~\cite{qwen3.5} are used as the evaluation metric.
}
  \subfloat[\bf \scriptsize Compression Method]{
    \resizebox{0.5\linewidth}{!}
    {
      \begin{tabular}{l|cccc}
        \bf Method & \bf Gemini$\uparrow$ & \bf GPT$\uparrow$ & \bf Qwen$\uparrow$ & \bf Avg. \\
        \midrule
        MLP + truncation & 0.179 & 0.153 & 0.176 & 0.169\\
        VAE expansion & 0.169 & 0.147 & 0.155 &  0.157\\
        \underline{MLP + MLP} & \underline{0.192} & \underline{0.170} &\underline{0.185} & \bf\underline{0.182}\\
      \end{tabular}
    }
  }
  \subfloat[\bf \scriptsize Modulation Method]
  {
    \resizebox{0.5\linewidth}{!}{
      \begin{tabular}{l|cccc}
        \bf Cross attention & \bf Gemini$\uparrow$ & \bf GPT$\uparrow$ & \bf Qwen$\uparrow$ & \bf Avg. \\
        \midrule
        Wo CA & 0.192 & 0.170 & 0.185& 0.182\\
        W Dual-Path CA & 0.227 & 0.185 & 0.229 & 0.214\\
        \underline{Dual-Path SAM}& \underline{0.242} & \underline{0.214} &\underline{0.238} & \bf\underline{0.231}\\
      \end{tabular}
    }
  } \\
  \vspace{1mm}
  \subfloat[\bf \scriptsize Training strategy]{
    \resizebox{0.5\linewidth}{!}
    {
      \begin{tabular}{l|cccc}
        \bf Train strategy & \bf Gemini$\uparrow$ & \bf GPT$\uparrow$ & \bf Qwen$\uparrow$ & \bf Avg. \\
        \midrule
        2-stage training & 0.336 & 0.256 & 0.283 & 0.291 \\
        \underline{Joint training} & \underline{0.540} & \underline{0.392} &\underline{0.503} & \bf\underline{0.478}\\
      \end{tabular}
    }
  }
\vspace{-3mm}
\label{tab:abl_main}
\end{table}
\paragraph{Different compression methods.}
Since our model leverages pre-trained weights, aligning the sequence length and feature dimensions of visual encoder to the pre-trained latent space is critical.
We explore three distinct compression and mapping strategies: (1) \textbf{MLP + truncation}: The feature dimension is mapped via an MLP, while the sequence length is directly truncated; (2) \textbf{VAE expansion}: The number of Transformer layers in the VAE is directly increased to naturally match dimensions within the latent space; (3) \textbf{MLP + MLP}: Both the sequence length and feature dimension are projected via MLPs.
As shown in Table~\ref{tab:abl_main}(a), the \textbf{MLP + MLP} strategy performs best (18.12\%).
We hypothesize that simple truncation discards critical edge information, while merely expanding VAE layers increases optimization difficulty.
Conversely, \textbf{dual MLP projection effectively preserves the semantic and spatial structure of the input visual prompt while maintaining pre-trained priors.}
\paragraph{Token gated cross attention.}
Next, we investigate modulation mechanisms to unify generation and editing, which exhibit distinct structural dependencies. 
We compare: (1) \textbf{Wo CA}: self-attention only; (2) \textbf{W Dual-Path CA}: dual-path cross-attention without adaptive gating; and (3) \textbf{Dual-Path SAM}: our proposed Dual-Path Spatially-Adaptive Modulation. 
Table~\ref{tab:abl_main}(b) shows that lacking cross-attention (Wo CA) yields the poorest results (18.12\%) due to insufficient utilization of source image structural priors. 
Adding cross-attention improves performance to 21.40\%, while Dual-Path SAM reaches 23.1\%. 
\textbf{This demonstrates that the adaptive gating mechanism dynamically balances content consistency and instruction adherence, optimizing performance within a unified framework.}
\paragraph{Joint training vs. two-stage training.}
Based on the optimal architecture described above, we evaluate data training strategies. We compare: (1) \textbf{Two stage training}: 100k steps on 3M samples (T2I, C2I, coarse-grained editing), followed by 140k steps on the remaining 2M samples; and (2) \textbf{Joint training}: mixing all 5M samples for 240k steps. As Table~\ref{tab:abl_main}(c) shows, joint training dominates with a \textbf{47.8\%} pass rate, far surpassing the two-stage approach (29.1\%). Two-stage training likely suffers from catastrophic forgetting across varying tasks. Instead, \textbf{joint training forces the model to simultaneously learn semantic generation, geometric transformation, and physical laws within a shared visual flow space, yielding stronger generalization and instruction adherence.}

\section{Conclusion}
We presented \modelname, a unified framework that redefines multimodal generation as a purely visual flow.
By embedding all modalities into a shared visual space and learning continuous transport between visual instruction and image states, \modelname achieves efficient and consistent generation across diverse tasks.
To support this paradigm, we introduced the large-scale \datasetname dataset, enabling cross-task generalization under a single visual interface.
\modelname achieves state-of-the-art performance among open-source models across all evaluated tasks and remains competitive with leading commercial systems in both automated and human evaluations.
These results highlight \modelname as a promising foundation for future vision-centric multimodal models that unify perception and generation under a single deterministic principle.
We believe this work marks a step toward closing the gap between visual understanding and creation within a continuous visual domain.
\clearpage  

\bibliographystyle{unsrtnat}
\bibliography{main}
\clearpage
\appendix
\renewcommand{\theHsection}{appendix.\Alph{section}}
\renewcommand{\theHsubsection}{appendix.\Alph{section}.\arabic{subsection}}
\section*{Appendix Overview}
\label{Appendov}
\vspace{-1em}
In the appendix, we provide additional information as listed
below:

\vspace{0.5em}
\noindent\rule{\textwidth}{1pt} 
\vspace{0.2em}

\newcommand{\tocsec}[2]{%
  \vspace{0.8em}\noindent\textbf{Appendix \ref{#1} \hspace{0.8em} #2} \hfill \textbf{p.~\pageref{#1}}\par
}
\newcommand{\tocsubsec}[2]{
  \vspace{0.3em}\noindent\hspace*{2.5em}\makebox[3em][l]{\ref{#1}} #2 \hfill \pageref{#1}\par
}

\tocsec{AppendA}{More experiment results}
\tocsubsec{subsec:error_analysis}{Error Analysis on human evaluation}
\tocsubsec{subsec:robust_test}{Robustness test}
\tocsubsec{subsec:vi_abla}{Visual instruction Ablation}
\tocsubsec{subsec:inter_agree}{Inter-evaluator agreement statistics}

\tocsec{AppendB}{More Benchmark Details}
\tocsubsec{subsec:leakage_pre}{Leakage prevention}
\tocsubsec{subsec:bench_analysis}{Data Analysis}

\tocsec{AppendC}{Additional Qualitative Examples}

\tocsec{AppendD}{More Dataset Details}
\tocsubsec{subsec:overview_visprompt}{Overview of \datasetname}
\tocsubsec{subsec:dataset_construct}{Data construction}
\tocsubsec{subsec:rendering_pipeline}{Visual Text Rendering Pipeline}
\tocsubsec{sec:quality_control}{Automated Quality Control and Filtering Pipeline}
\tocsubsec{subsec:dataset_composition}{Dataset Composition and Detailed Statistics}

\tocsec{AppendE}{Limitations and future work}

\tocsec{AppendF}{More evaluation details}
\tocsubsec{vlm_eval}{VLM evaluation}
\tocsubsec{subsec:eval_details}{Quantitative and human evaluation}

\tocsec{AppendG}{Model details}
\tocsubsec{subsec:param_breakdown}{Parameter breakdown}
\tocsubsec{subsec:loss_func}{Loss function for image in-image out generation}
\tocsubsec{hyperparms_abla}{Hyperparameter ablation}
\tocsubsec{exp_details}{Experimental details}

\vspace{0.8em}
\noindent\rule{\textwidth}{1pt} 
\vspace{1em}
\clearpage
\section{More experiment results}
\label{AppendA}
\subsection{Error Analysis on human evaluation}
\label{subsec:error_analysis}
To complement our automated VLM-based evaluation and gain deeper insights into the failure modes of our model, we conducted a rigorous human evaluation. 
Ten independent expert evaluators independently assessed, and subsequently cross-checked, a stratified random subset of 250 generated samples across the benchmark (25 samples per subset).
Note that the methodological difference between our human evaluation and the VLM evaluation: while the VLM assigns a continuous score (1-5) for every sample across all four dimensions, the human evaluators adopted a strict visual-inspection approach. \begin{wrapfigure}{r}{0.45\textwidth}
    \centering
    \includegraphics[width=0.3\textwidth]{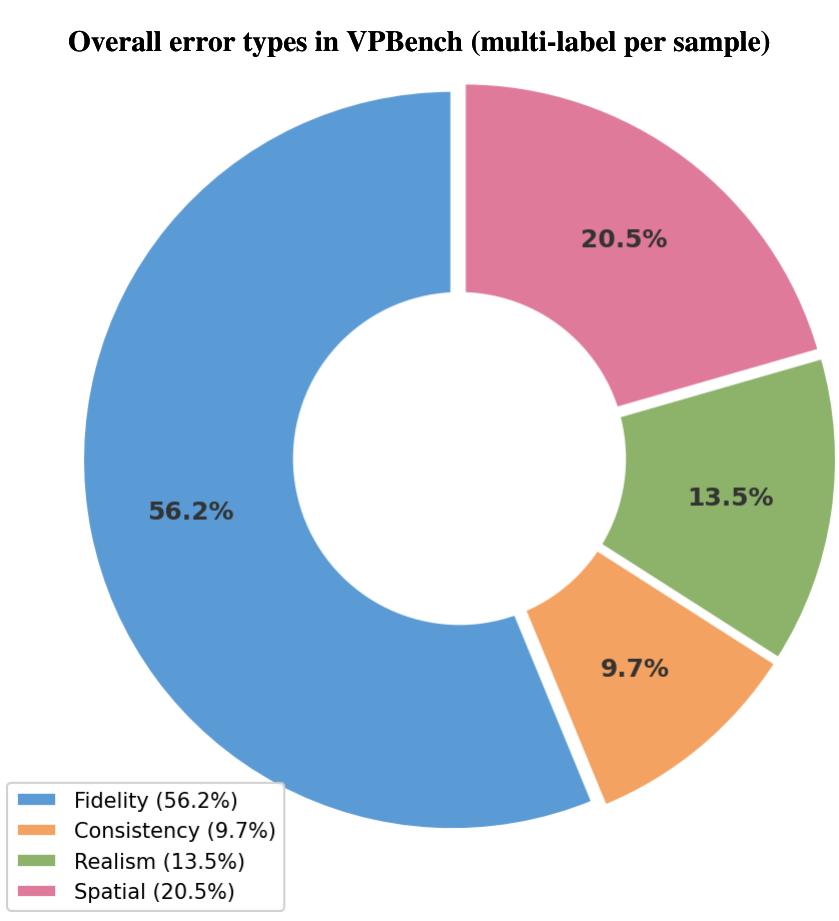}
    \caption{\textbf{Overall} error types in \benchname. For brevity in the figures, the labels Fidelity, Spatial, Realism, and Consistency correspond strictly to Instruction Fidelity, Spatial Precision, Visual Realism, and Content Consistency, respectively.}
    \label{fig:fail_dist_global}
\end{wrapfigure}
Specifically, evaluators first made a binary judgment on whether the generated image was qualified. 
If a sample was deemed a ``FAIL,'' the evaluator then recorded which specific dimensions contributed to the failure, allowing for multi-label tagging per sample.
Please note that for brevity in the figures, the labels Fidelity, Spatial, Realism, and Consistency correspond strictly to Instruction Fidelity, Spatial Precision, Visual Realism, and Content Consistency, respectively.

As illustrated in Figure~\ref{fig:fail_dist_global}, the global error distribution across the entire VP-Bench reveals that Instruction Fidelity is the most prominent bottleneck, accounting for $56.2\%$ of all tagged errors. 
This indicates that fully capturing the nuanced semantics of complex visual instructions remains the primary challenge.
Spatial Precision constitutes the second largest error source at $20.5\%$, followed by Visual Realism ($13.5\%$) and Content Consistency ($9.7\%$).

\begin{figure*}[htbp]
    \centering
    \includegraphics[width=0.92\textwidth]{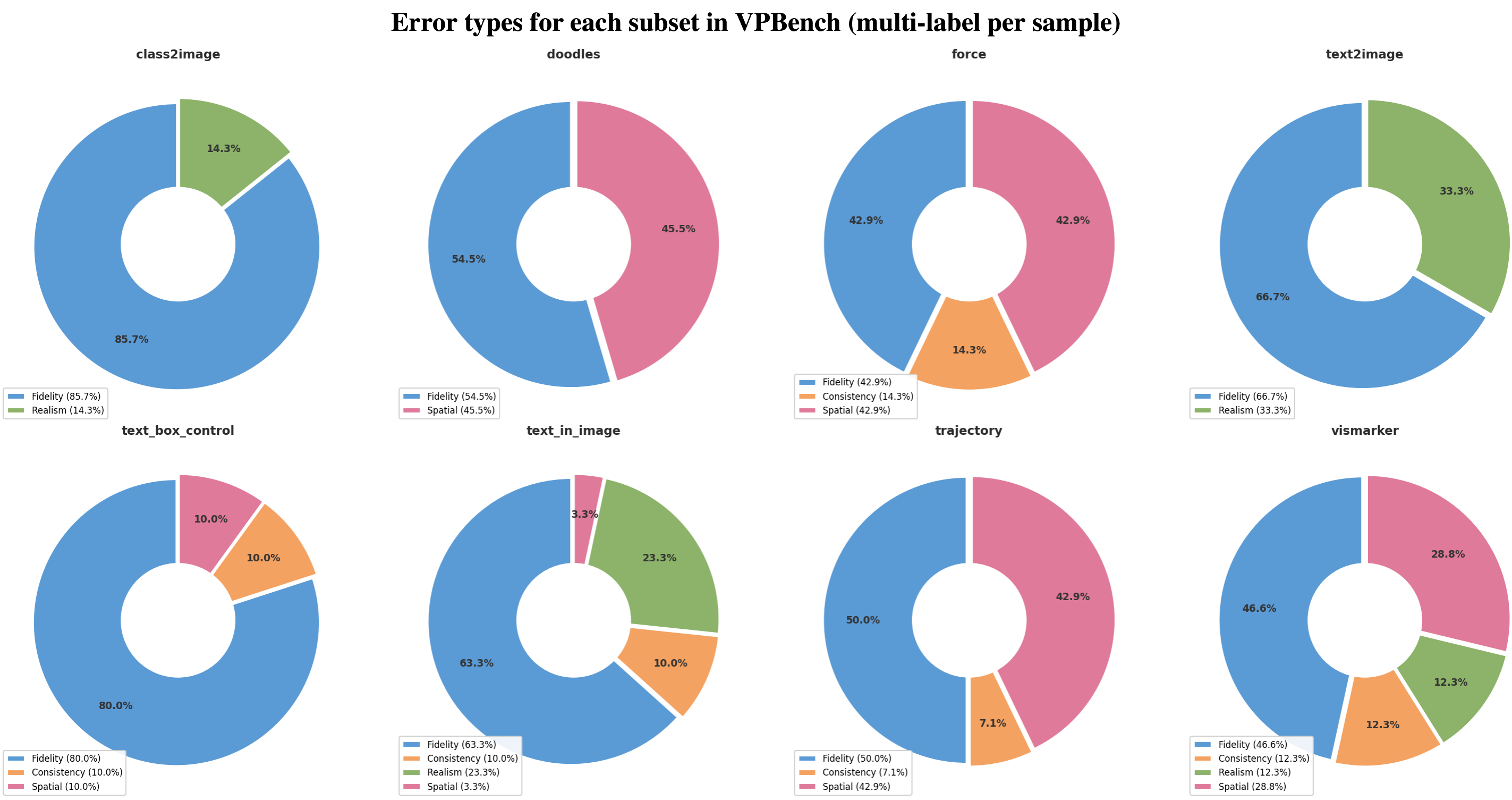}
    \caption{\textbf{Error types for each subset in \benchname.}}
    \vspace{-5mm}
    \label{fig:fail_dist_local}
\end{figure*}

A more granular breakdown of error types by subset category is presented in Figure~\ref{fig:fail_dist_local}. 
The distribution of errors varies significantly depending on the nature of the specific task.
For instance, in semantics-driven tasks that require strict adherence to explicit textual concepts and precise content generation, such as class-to-image and text bbox control, Instruction Fidelity errors overwhelmingly dominate, taking up $85.7\%$ and $80.0\%$ respectively. 
Conversely, in tasks requiring strict spatial grounding and geometric reasoning—such as doodles, force, trajectory, and vismarker—the proportion of Spatial Precision errors significantly increases, reaching up to $45.5\%$ in doodle-guided tasks. 
Additionally, Visual Realism emerges as a more noticeable issue in from-scratch generation tasks like text-to-image ($33.3\%$). 

\subsection{Robustness test}
\label{subsec:robust_test}
\begin{figure*}[htbp]
    \centering
    \includegraphics[width=0.96\textwidth]{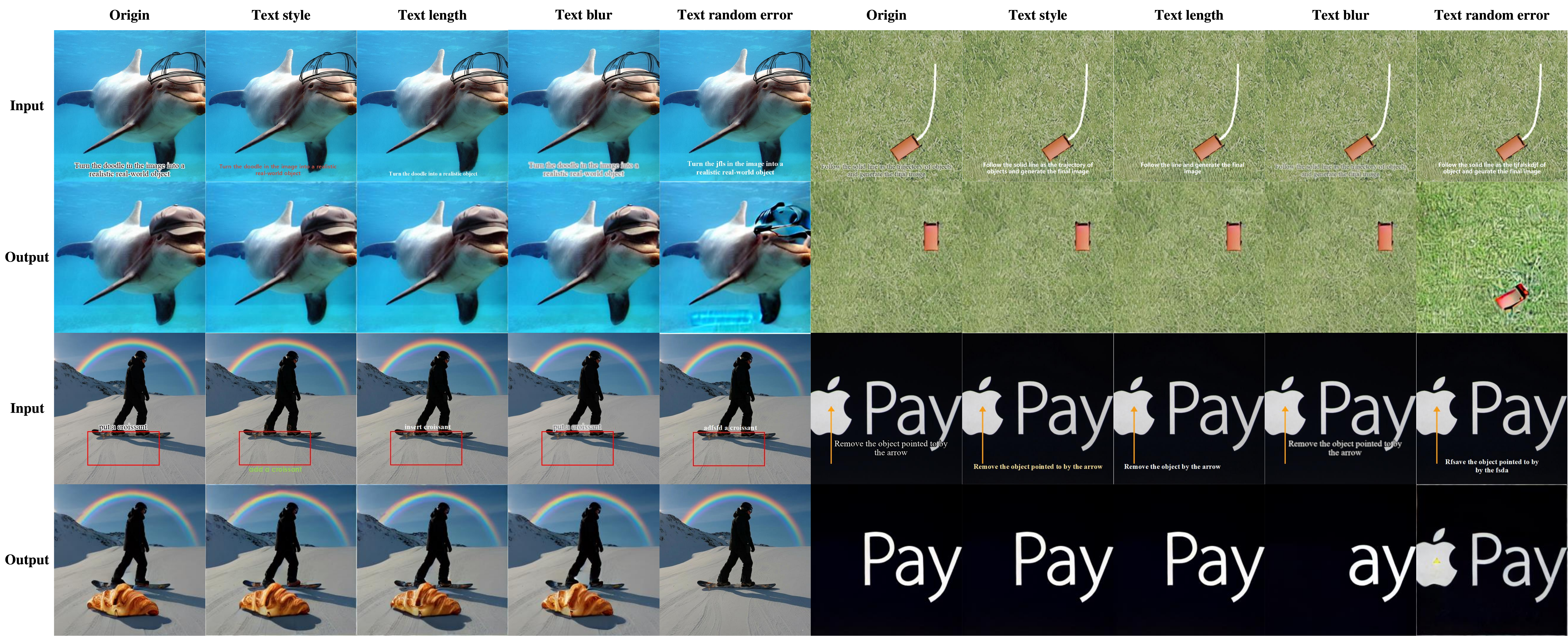}
    \caption{\textbf{Qualitative results of the robustness analysis against visual instruction perturbations.} We evaluate the model's generation stability under five distinct conditions: the original unmodified input, variations in text style (e.g., size, color, font, and layout), changes in text length, text blurring, and random text corruption.}
    \vspace{-3mm}
    \label{fig:robust}
\end{figure*}
To evaluate the stability and reliability of our model in real-world scenarios, we conduct comprehensive robustness tests under various challenging conditions. 

First, we assess the resilience of model to perturbations in the textual components of the visual instructions. 
As shown in Figure~\ref{fig:robust}, we apply four distinct types of interference to the original input: changes in text style (including size, color, font, and spatial layout), variations in text length, severe text blurring, and random text corruption (replacing valid instructions with meaningless or incorrect text). 
The generation results demonstrate that our model maintains highly stable and accurate performance across style changes, length variations, and strong blurring. 
\textbf{\textcolor{red}{This indicates that the model effectively extracts the underlying semantic intent rather than simply memorizing superficial formatting.}} 
However, as expected, the performance of model significantly deteriorates when subjected to random text errors. 
\textbf{This failure case actually serves as positive confirmation that our model strictly follows the explicit semantic guidance provided within the image, rather than hallucinating edits based on visual context alone.}

\begin{figure*}[htbp]
    \centering
    \includegraphics[width=0.6\textwidth]{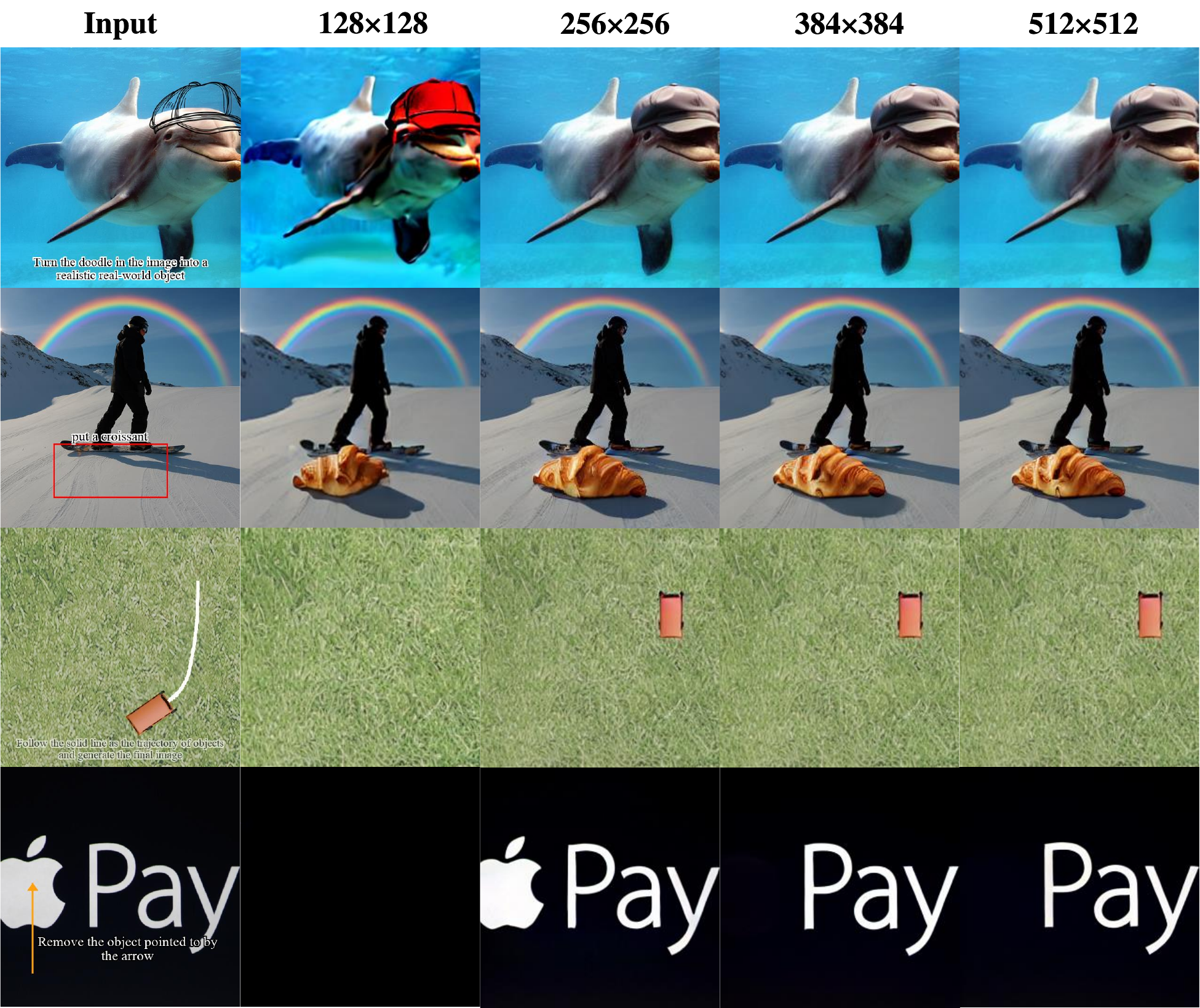}
    \caption{\textbf{Generation results across different input image resolutions.}}
    \label{fig:robust_resolu}
\end{figure*}
Furthermore, we evaluate the robustness of model across different resolutions. We test the generation quality using input images scaled to $128 \times 128$, $256 \times 256$, $384 \times 384$, and $512 \times 512$ pixels.
As illustrated in Figure~\ref{fig:robust_resolu}, the model is largely capable of producing high-quality generation at resolutions of $256 \times 256$ and above, despite occasional failure cases.
A noticeable drop in visual fidelity and instruction-following capability is only observed at the extreme low resolution of $128 \times 128$. 
This is primarily because extreme downsampling severely compresses the visual markers and text, making them illegible for accurate feature extraction.

\subsection{Visual instruction Ablation}
\label{subsec:vi_abla}
To thoroughly investigate the individual contributions of textual and visual components within our unified visual instructions, we conducted an ablation study across four representative samples.
Specifically, we evaluated the generation performance under four distinct input configurations: (1) \textbf{Blank}, where both text and visual prompts are removed, leaving only the original source image; (2) \textbf{Text}, where visual prompts are removed; (3) \textbf{Visual prompt}, where textual instructions are removed; and (4) \textbf{Text + visual prompt}, representing the complete visual instruction.
\begin{figure*}[htbp]
    \centering
    \includegraphics[width=\textwidth]{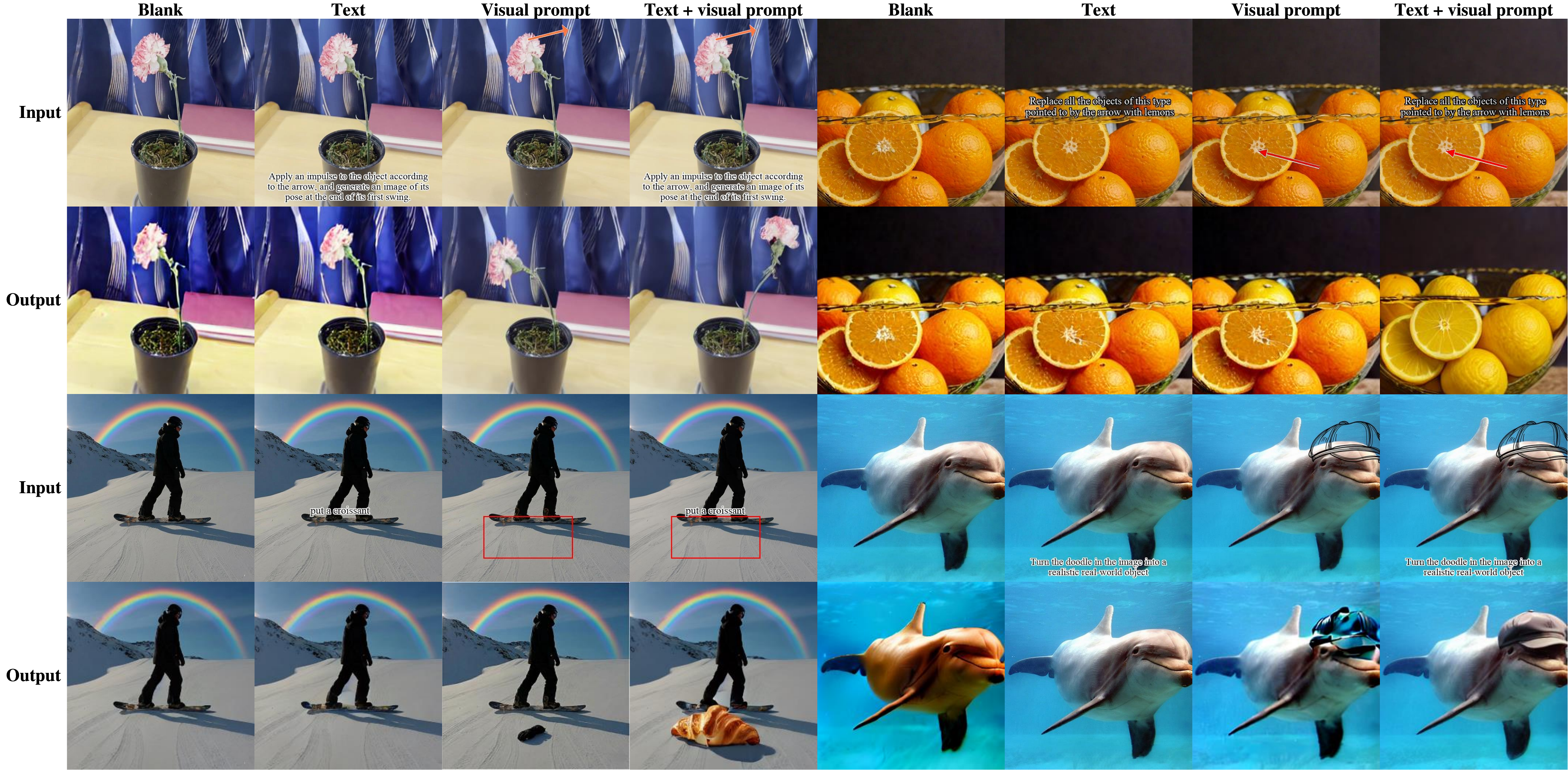}
    \caption{\textbf{Qualitative results of the visual instruction ablation study.} We compare generation outcomes across four input configurations: Blank (original image only), Text only, Visual prompt only, and the complete Text \& visual prompt. }
    \label{fig:vi_abla}
\end{figure*}
As illustrated in Figure~\ref{fig:vi_abla}, the results clearly demonstrate the necessity of combining both instructional elements within the unified image canvas. 
When provided with the ``Blank'' image, the model predictably makes no modifications. 
In the ``Text only'' setting, the model primarily acts to erase the overlaid text but fails to execute the core editing intent, as it lacks precise spatial and operational grounding. 
Conversely, the ``Visual prompt only'' setting leads to chaotic and disordered generation results.
\textbf{This strongly proves that the in-image text is indispensable, serving as a lightweight semantic trigger that assigns explicit functional meaning to the visual prompts. }
Finally, \textbf{only the complete ``Text + visual prompt'' configuration successfully and consistently generates the correct, high-quality target images, confirming that the synergy between rendered textual semantics and graphical spatial cues is essential for accurate visual instruction following.}
\subsection{Inter-evaluator agreement statistics}
\label{subsec:inter_agree}
As shown in Table~\ref{tab:inter_evaluator_agreement}, despite absolute score variations caused by different VLM evaluators' judging preferences and biases, the inter-evaluator agreement remains reasonably strong. 
Specifically, the \textbf{Kendall's $W$ score of 0.7422}, which measures the consistency of ordinal model rankings across evaluators, indicates that different evaluators produce largely aligned relative rankings among the compared models.
Meanwhile, the \textbf{Fleiss' $\kappa$ score of 0.6132}, which reflects the agreement on binary pass/fail judgments, further suggests substantial consensus at the instance level. 
These results demonstrate that although individual evaluators may assign different absolute success rates, the overall comparative conclusions are stable and not dominated by evaluator-specific biases.
\begin{table}[htbp]
\centering
\caption{\textbf{Inter-evaluator Agreement.}}
\label{tab:inter_evaluator_agreement}
\small
\setlength{\tabcolsep}{4pt}
\renewcommand{\arraystretch}{1.0}
\begin{tabular}{lcc}
\toprule
\textbf{Model} & \textbf{Fleiss' $\kappa$} & \textbf{Kendall's $W$} \\
\midrule
FlowInOne      & 0.6537 & 0.7935 \\
FLUX.1 Kontext & 0.4540 & 0.5189 \\
Nano Banana    & 0.4537 & 0.5251 \\
OmniGen2       & 0.5361 & 0.5082 \\
Qwen-IE-2509   & 0.4515 & 0.6115 \\
\midrule
\textbf{Overall} & \textbf{0.6132} & \textbf{0.7422} \\
\bottomrule
\end{tabular}
\end{table}
\section{More Benchmark Details}
\label{AppendB}
\begin{table}[htbp]
\centering
\caption{\textbf{Detailed statistics of the human-curated \benchname}. To provide a structured overview, highly fine-grained editing sub-categories are conceptually grouped into higher-level semantic macro-categories.}
\label{tab:benchmark_statistics}
\resizebox{0.9\linewidth}{!}{
\begin{tabular}{@{}llcc@{}}
\toprule
\textbf{Top-Level Task} & \textbf{Sub-Category / Grouping} & \textbf{Pairs} & \textbf{Total} \\ \midrule
Class-to-Image & - & - & 100 \\ \midrule
Doodles Editing & - & - & 50 \\ \midrule
\multirow{2}{*}{Force Understanding} 
 & Point Force & 100 & \multirow{2}{*}{150} \\
 & Global Force & 50 &  \\ \midrule
Text-to-Image & - & - & 50 \\ \midrule
Text BBox Control & - & - & 50 \\ \midrule
\multirow{5}{*}{Text-in-Image Editing} 
 & Semantic Operations (Add, Remove, etc.) & 65 & \multirow{5}{*}{290} \\
 & Attribute \& Environment Modifications & 70 &  \\
 & Style \& Artistic Transfer & 20 &  \\
 & Structural Tasks (Condition-to-Image) & 65 &  \\
 & Complex Spatial \& Reasoning & 70 &  \\ \midrule
Trajectory Understanding & - & - & 50 \\ \midrule
\multirow{3}{*}{Visual Marker Editing} 
 & Removal Operations & 40 & \multirow{3}{*}{320} \\
 & Replacement \& Swapping & 120 &  \\
 & Attribute \& Local Changes & 160 &  \\ \midrule \midrule
\textbf{Total Benchmark Size} & & \multicolumn{2}{c}{\textbf{1,060 Pairs}} \\ \bottomrule
\end{tabular}
}
\end{table}
\subsection{Leakage prevention}
\label{subsec:leakage_pre}
Since \benchname is derived via random sampling from our eight task categories, we implemented a strict two-fold leakage prevention protocol to ensure a genuine zero-shot evaluation and defend against data contamination:
\begin{itemize}
    \item \textbf{Root-Image Level Partitioning:} Instead of splitting the dataset at the final instruction-pair level, the split is strictly executed based on the underlying \textit{root images} (the unedited base canvases). 
    This ensures the model has never observed the semantic backgrounds, objects, or spatial layouts of the benchmark during optimization.
    \item \textbf{Variant and Augmentation Exclusion via Visual Deduplication:} To technically enforce this mutual exclusivity, we employed a rigorous visual feature deduplication pipeline. 
    Specifically, we extracted deep visual embeddings via CLIP~\cite{pmlr-v139-radford21a} for all root images sampled for \benchname.
    We then computed the cosine similarity against the entire 5M training pool. 
    Any training pair exhibiting a visual similarity score above a highly conservative threshold was aggressively discarded. 
    This feature-level filtering mathematically guarantees that no differently instructed pairs, cropped variants, or intermediate edits originating from the benchmark's base canvases remain in the training phase.
\end{itemize}

\subsection{Data Analysis}
\label{subsec:bench_analysis}
\begin{figure*}[t]
    \centering
    
    \begin{minipage}{0.4\textwidth}
        \centering
        \includegraphics[width=\linewidth]{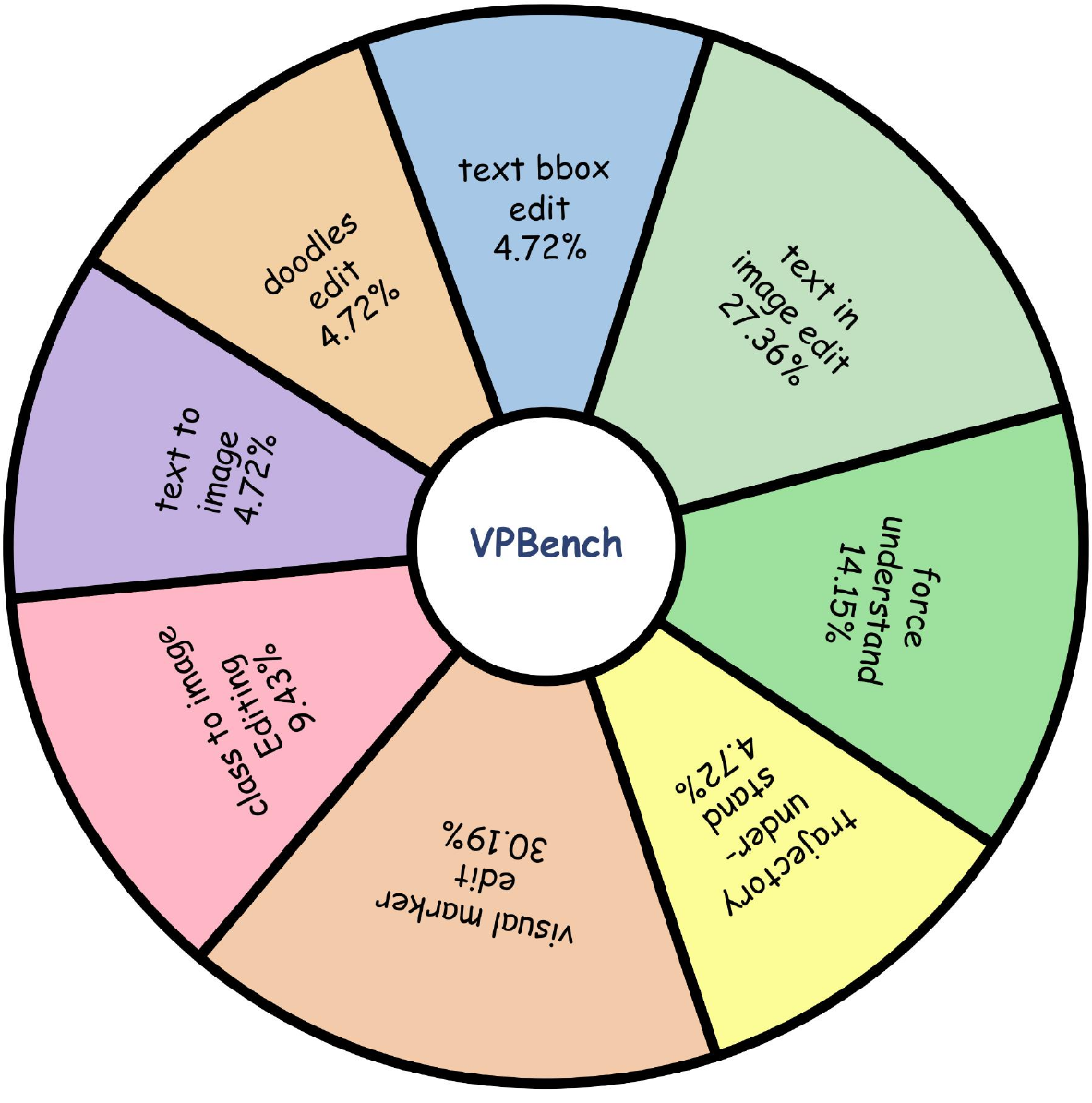}
        \caption{\textbf{Data distribution across the eight distinct subsets within \benchname.}\benchname is a comprehensive benchmark that comprises eight distinct data types.}
        \label{fig:dataset_ratio}
    \end{minipage}
    \hfill 
    \begin{minipage}{0.5\textwidth}
        \centering
        \includegraphics[width=\linewidth]{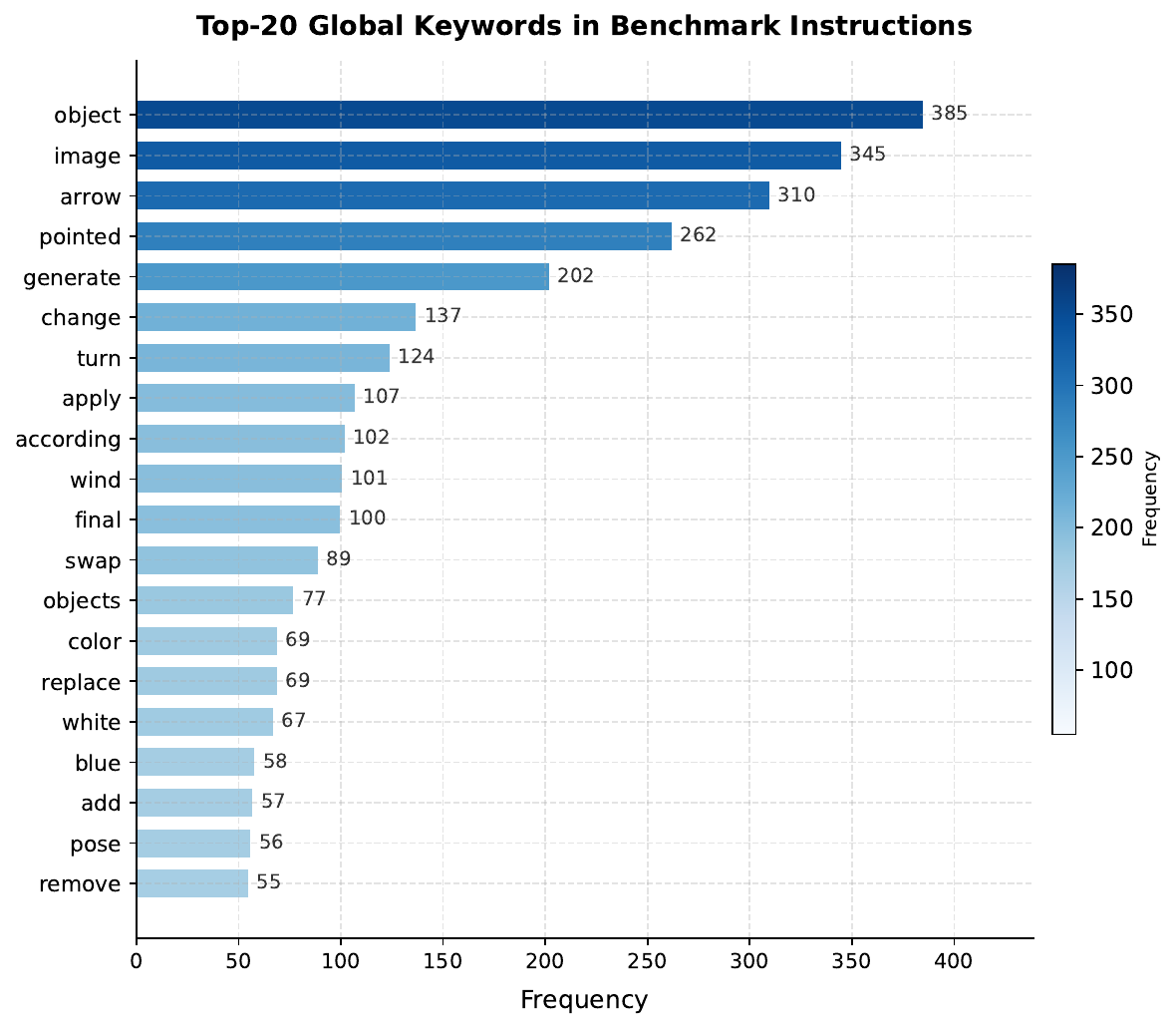}
        \caption{\textbf{Top-20 global keywords in \benchname instructions.} The distribution reveals a dominance of operational verbs (e.g., ``change'', ``turn'') and visual pointers (e.g., ``arrow''), highlighting the minimal text dependency of our visual-centric evaluation paradigm.}
        \label{fig:keyword_dist}
    \end{minipage}
    
    \vspace{-4mm}
\end{figure*}

\noindent\textbf{Overview of \benchname.}
As summarized in Table~\ref{tab:benchmark_statistics} and Figure~\ref{fig:dataset_ratio}, \benchname comprises a total of 1,060 meticulously curated image pairs spanning eight distinct task categories.
The dataset is heavily anchored by two core visual instruction tasks: \textit{Visual Marker Editing} (320 pairs, 30.19\%) and \textit{Text-in-Image Editing} (290 pairs, 27.36\%). 
Together, these constitute over 57\% of the benchmark, providing highly fine-grained evaluations across semantic operations, structural tasks, and complex Spatial Reasoning. 
\begin{table}[htbp]
\centering
\caption{\textbf{Textual prompt statistics across different task categories in \benchname.} The uniquely low average word counts prove the benchmark's reliance on explicit visual markers rather than lengthy textual descriptions.}
\label{tab:text_statistics}
\renewcommand{\arraystretch}{1.15} 
\setlength{\tabcolsep}{4.5mm} 
\resizebox{\linewidth}{!}{
\begin{tabular}{@{}lcccc@{}}
\toprule
\textbf{Task Category} & \textbf{Pairs} & \textbf{Avg. Words} & \textbf{Word Range} & \textbf{Avg. Chars} \\ \midrule
Class-to-Image & 100 & 3.0 & 1 -- 6 & 22.4 \\
Text BBox Control & 43 & 3.5 & 3 -- 5 & 16.0 \\ \midrule
Doodles Editing & 43 & 12.0 & 12 -- 12 & 63.0 \\
Trajectory Understanding & 50 & 14.0 & 14 -- 14 & 79.0 \\
Visual Marker Editing & 312 & 14.1 & 3 -- 32 & 74.3 \\
Force Understanding & 150 & 17.7 & 13 -- 24 & 96.0 \\ 
Text-in-Image Editing & 292 & 17.9 & 2 -- 44 & 105.0 \\ \midrule \midrule
Text-to-Image (Baseline) & 50 & 23.7 & 8 -- 46 & 137.1 \\ \bottomrule
\end{tabular}
}
\end{table}
Beyond general editing, \benchname is strategically structured to evaluate specific advanced capabilities.
To rigorously test physics-aware reasoning, we specifically include \textit{Force Understanding} (150 pairs, 14.15\%) and \textit{Trajectory Understanding} (50 pairs, 4.72\%). 
Additionally, the benchmark evaluates explicit spatial layout and sketch-based control through \textit{Text BBox Control} (50 pairs, 4.72\%) and \textit{Doodles Editing} (50 pairs, 4.72\%). 
Finally, to ensure a comprehensive evaluation spectrum, we incorporate \textit{Class-to-Image} (100 pairs, 9.43\%) and the baseline \textit{Text-to-Image} (50 pairs, 4.72\%) purely as foundational tasks.
This precise proportional distribution ensures that the benchmark prioritizes complex visual grounding and physical reasoning over conventional text-driven generation.
We deliberately designed this proportional distribution to ensure that subsets driven by explicit visual prompts heavily dominate the benchmark. 
This structural emphasis highlights our core objective: rigorously evaluating a model's capacity for genuine visual-centric instruction following, thereby preventing models from bypassing visual grounding through conventional text-driven priors.

\noindent\textbf{Additional statistics.}
A core motivation of \benchname is to shift the cognitive load from complex, dense textual descriptions to intuitive, explicit visual instructions. 
To quantitatively demonstrate that our benchmark evaluates genuine visual reasoning rather than textual comprehension, we analyze both the linguistic length and the semantic distribution of the textual prompts extracted in all images.
First, as detailed in Table~\ref{tab:text_statistics}, the overall textual requirement is extremely lightweight. 
TTasks driven by straightforward semantic categories (e.g., \textit{Class-to-Image}) or explicit spatial bounding boxes (e.g., \textit{Text BBox Control}) utilize highly concise textual tags, averaging just 3.0 and 3.5 words, respectively. 
Even for complex physics-aware reasoning (\textit{Trajectory Understanding} and \textit{Force Understanding}) and intricate manipulation (\textit{Visual Marker Editing}), the text remains remarkably brief (averaging between 14 to 18 words). 
Because such a brief text alone is insufficient to describe complex spatial configurations, the model is forced to extract the precise operational intent directly from the spatial geometry of the rendered boxes, doodles, and superimposed arrows. 
In contrast, the traditional \textit{Text-to-Image} baseline relies on significantly longer descriptive prompts (averaging 23.7 words). 

Furthermore, the semantic composition of the instructions corroborates this visual reliance.
As illustrated in Figure~\ref{fig:keyword_dist}, the top-20 global keywords diverge sharply from traditional generation prompts.
Instead of dense descriptive adjectives, the vocabulary is dominated by abstract references (e.g., ``object'', ``image''), explicit visual pointers (e.g., ``arrow'', ``pointed''), and operational verbs (e.g., ``change'', ``turn'', ``swap''). 
This stark statistical and linguistic contrast robustly validates that \benchname minimizes text dependency, utilizing text merely as a lightweight operational trigger while rigorously testing ``image-in, image-out'' visual instruction following.
\section{Additional Qualitative Examples}
\label{AppendC}
\begin{figure*}[htbp]
    \centering
    \includegraphics[width=\textwidth]{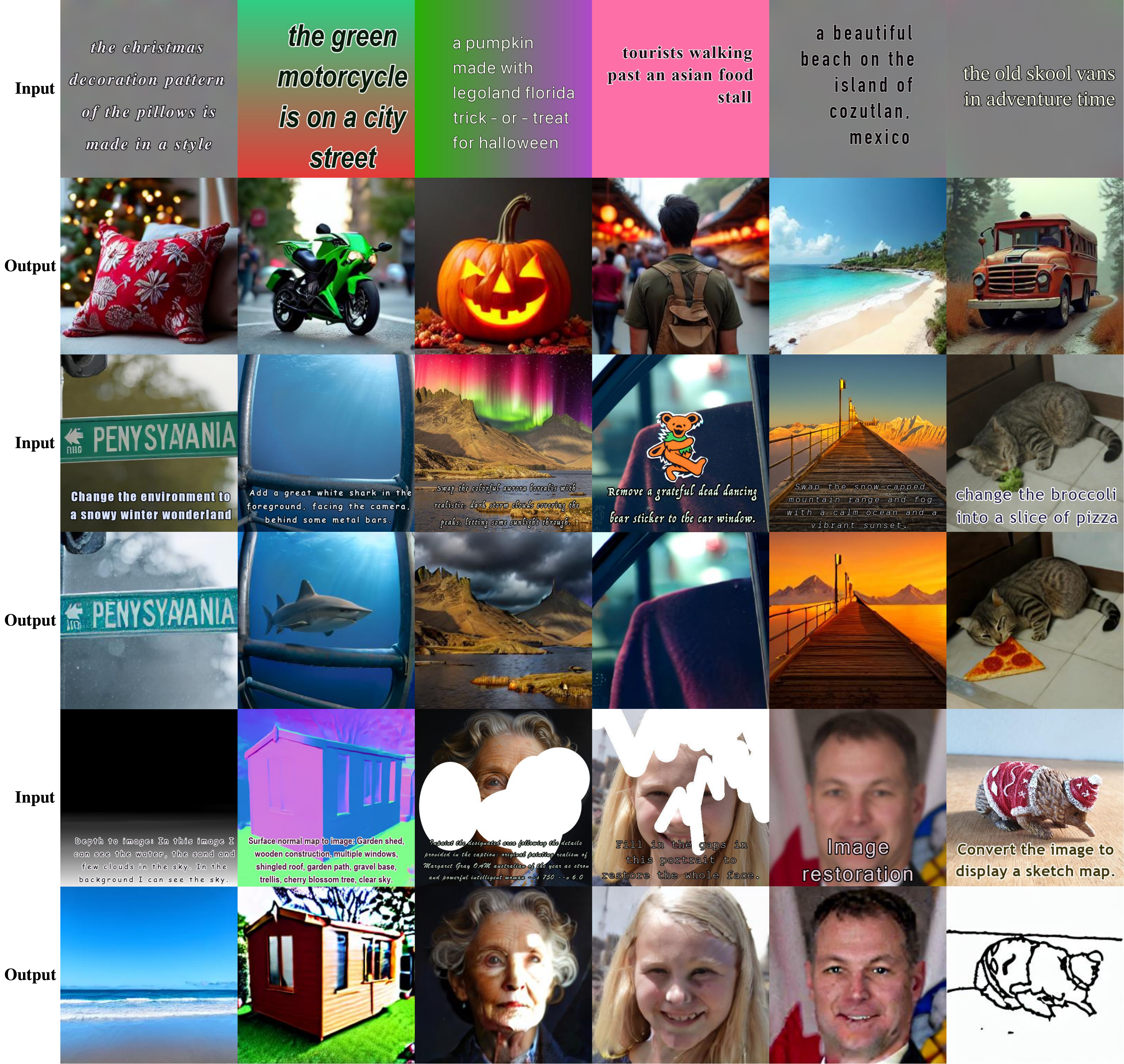}
    \caption{\textbf{Additional qualitative results for text-to-image generation and text-in-image editing tasks.}}
    \label{fig:quan1}
\end{figure*}
\begin{figure*}[htbp]
    \centering
    \includegraphics[width=\textwidth]{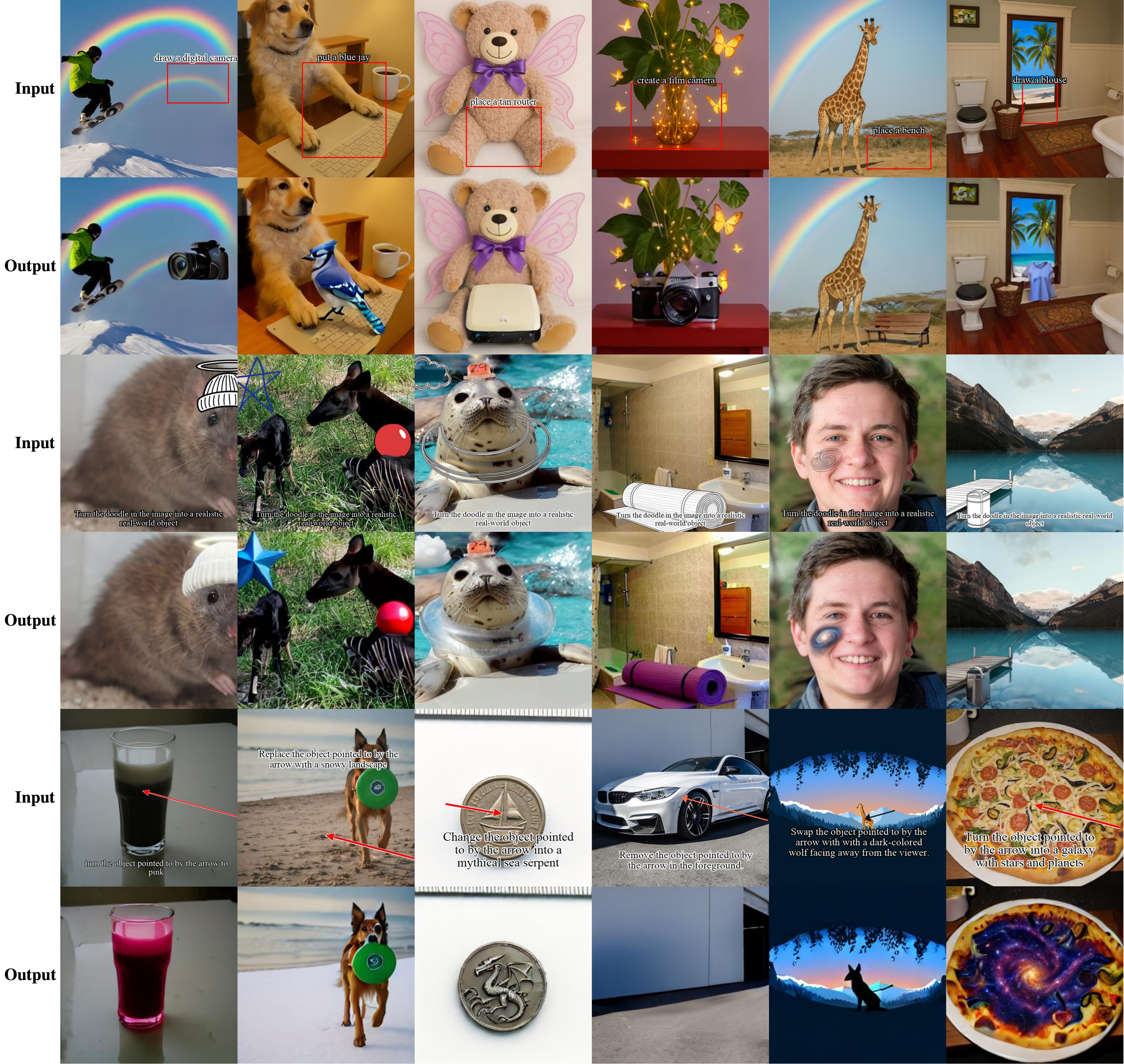}
    \caption{\textbf{Additional qualitative results for text bounding box (bbox) editing, doodle-guided editing, and visual marker-based editing.}}
    \label{fig:quan2}
\end{figure*}
\begin{figure*}[htbp]
    \centering
    \includegraphics[width=\textwidth]{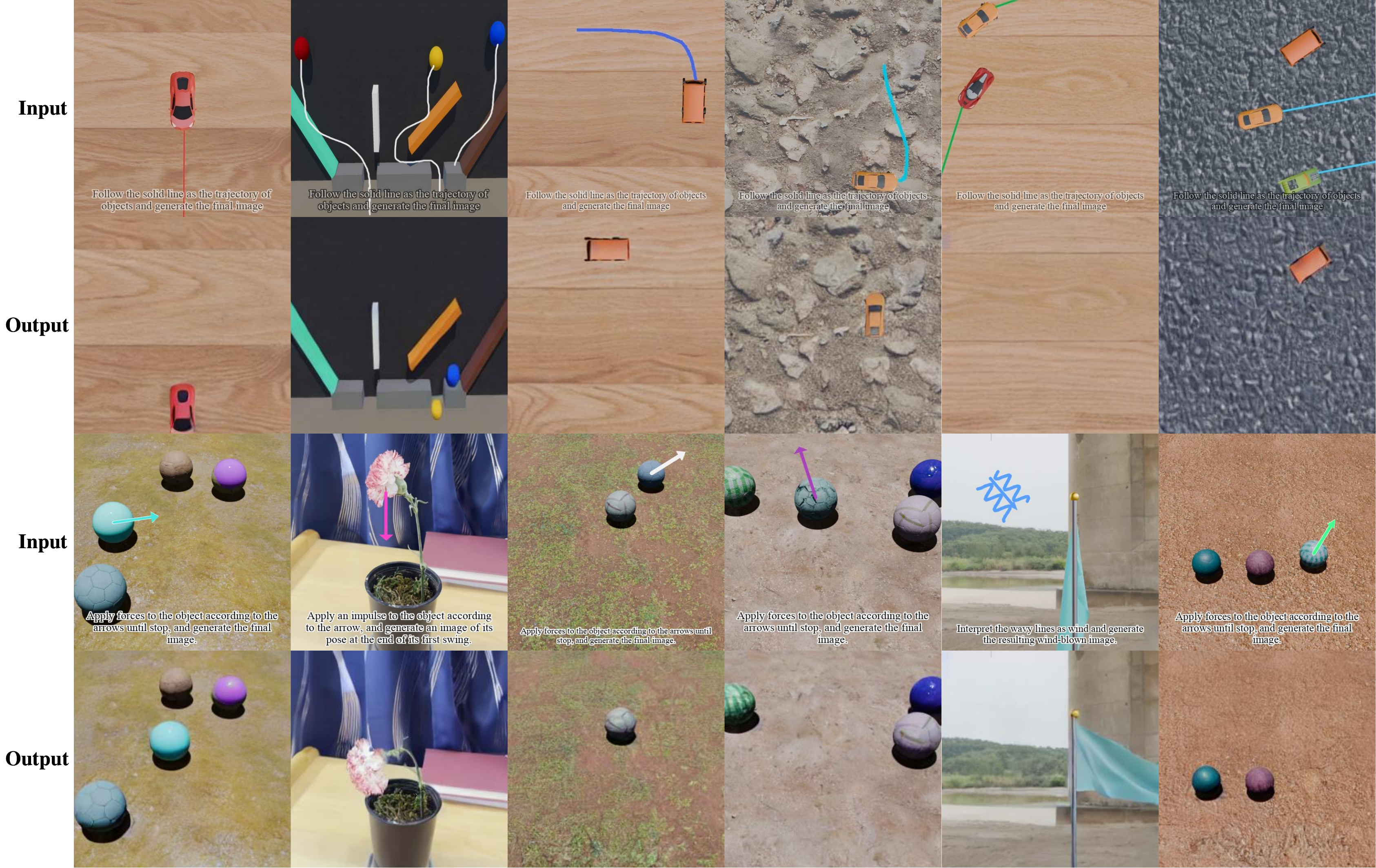}
    \caption{\textbf{Additional qualitative results demonstrating physical force understanding and trajectory understanding.}}
    \label{fig:quan3}
\end{figure*}

In this section, we provide additional qualitative examples to further demonstrate the versatile generation capabilities of our model across various visual instruction categories. Specifically, we group these supplementary results into three main aspects:
(1) Figure~\ref{fig:quan1} showcases extended results for text-to-image generation and text-in-image editing tasks. 
(2) Figure~\ref{fig:quan2} provides further visual examples focusing on text bounding box (bbox) editing, doodle-guided editing, and visual marker-based editing. 
(3) Figure~\ref{fig:quan3} illustrates additional generation outcomes that emphasize the model's capacity for physical force understanding and trajectory understanding.

\section{More Dataset Details}
\label{AppendD}
\subsection{Overview of \datasetname}
\label{subsec:overview_visprompt}
To support the training of \modelname under a purely vision-centric paradigm, we constructed \textbf{\datasetname}, a meticulously curated large-scale dataset comprising approximately 5 million pairs of visual instructions.
Unlike traditional multimodal datasets that rely on disjointed textual captions, \datasetname unifies diverse control signals by explicitly rendering task-specific instructional elements---such as text, spatial layouts, or physical constraints---directly onto the input image canvas.

Table~\ref{tab:dataset_overview} provides a comprehensive statistical breakdown of \datasetname. To ensure \modelname acquires a versatile and generalizable visual reasoning capability, we scale the dataset across eight distinct fine-grained task categories. These categories can be conceptually grouped into three major multimodal capabilities:
\begin{itemize}
    \item \textbf{Fundamental Generation (\textit{Text to Image, Class to Image}):} Constructed via direct rendering of textual prompts and class labels from massive datasets, establishing the basic semantic-to-visual mapping directly on the input images.
    \item \textbf{Unified Image Editing (\textit{Text in Image Edit, Text Bounding Box Edit, Visual Marker Edit, Doodles Edit}):} This overarching category spans a broad spectrum of manipulations. 
    It ranges from general operational intents (e.g., attribute changes, inpainting, and condition-to-image tasks extracted from large-scale datasets) to precise, spatially-constrained modifications. 
    Notably, the subsets requiring explicit geometric priors and implicit visual cues (BBox, Visual Markers, and Doodles) are specifically synthesized and heavily curated through advanced MLLMs (e.g., Qwen3-VL) and rigorous manual inspection to guarantee high spatial fidelity.
    \item \textbf{Physics Understanding (\textit{Trajectory Understanding, Force Understanding}):} The most challenging subset, pushing the boundary of physics-aware image-in, image-out generation. It incorporates manually annotated trajectory lines and explicit force arrows, visualizing precise object dynamics.
\end{itemize}

As illustrated in Table~\ref{tab:dataset_overview}, our rigorous construction and filtering pipeline, spanning from raw data extraction to multi-stage MLLM synthesis and manual annotation—ensures that the model learns accurate spatial-visual logic rather than exploiting noisy dataset correlations.
\begin{table*}[htbp]
\centering
\caption{\textbf{Comprehensive Overview of the VisPrompt-5M Dataset.} We detail the task categories, original data sources, data scale before and after the filtering pipeline, construction methodology, and the explicitly embedded visual instruction formats.}
\label{tab:dataset_overview}
\resizebox{\textwidth}{!}{%
\begin{tabular}{@{}ll r@{\hspace{1.5em}}r @{\hspace{2em}}ll@{}}
\toprule
\textbf{Task Category} & \textbf{Raw Source(s)} & \textbf{Raw Size} & \textbf{Retained Size} & \textbf{Construction Pipeline} & \textbf{Visual Instruction Format} \\ \midrule
\bf Text to Image & text-to-image-2M~\cite{zk_2024} & 2.26M & 2.26M & Direct Rendering & Text \\
\bf Class to Image & ImageNet (Subset)~\cite{Russakovsky2014ImageNetLS} & 2M & 860K & Direct Rendering & Text (Class Labels) \\ \midrule
\multirow{4}{*}{\bf Text in Image Edit} & GPT-Image-Edit~\cite{wang2025gptimageedit15mmillionscalegptgeneratedimage} & 1.5M & 1.04M & \multirow{3}{*}{Extracted \& Filtered} & \multirow{4}{*}{Text} \\
 & Pico-Banana~\cite{qian2025picobanana400klargescaledatasettextguided} & 400K & 10K & & \\
 & UnicEdit~\cite{ye2025unicedit} & 2M & 585K & & \\
 & PixWizard~\cite{lin2024pixwizard} & 500K & 315K & Extracted & \\ \midrule
\bf Text \& BBox Edit & GPT-Image-Edit subset~\cite{wang2025gptimageedit15mmillionscalegptgeneratedimage} & 45K & 24K & Synthetic Editing (MLLM guided) & Text + Bounding Box \\ \midrule
\bf Visual Marker Edit & UltraEdit subset~\cite{zhao2024ultraedit} and OmniEdit~\cite{wei2024omniedit} & 400K & 250K & Synthetic Editing \& Filtered & Text + Arrow Marker \\ \midrule
\bf Doodles Edit & Web Crawled Images & 5K & 1K & Two-stage Synthesis \& Manual & Text + Doodle \\ \midrule
\bf Trajectory Understanding & Blender-rendered Videos & - & 1.5K & Manual Annotation & Text + Trajectory Line \\ \midrule
\bf Force Understanding & Force Prompting Dataset~\cite{gillman2025forcepromptingvideogeneration} & 36K & 32K & Keyframe Extracted \& Annotated & Text + Force Arrow \\ \bottomrule
\end{tabular}%
}
\end{table*}

\subsection{Data construction.}
\label{subsec:dataset_construct}
\paragraph{Fundamental Generation}
To align with our unified image-in, image-out paradigm, we must convert traditional text-image pairs into unified image-image pairs.
For the \textbf{text-to-image} task, we directly render the corresponding textual prompts onto a blank input canvas. 
To ensure the model acquires robust visual text comprehension rather than overfitting to specific typographical layouts, we introduce extensive data augmentation during the rendering process. 
Specifically, the font style, font size, font color, text position, and the background color of the canvas are all randomly sampled.
We also enforce a strict boundary check to ensure that all generated text remains entirely within the canvas limits.
Through this automated pipeline, the 2M text-image pairs from text-to-image-2M~\cite{zk_2024} are seamlessly transformed into image-image pairs.

Similarly, for the \textbf{class-to-image} generation task, we utilize a high-quality subset of ImageNet~\cite{Russakovsky2014ImageNetLS}. 
The discrete class labels (e.g., ``golden retriever'') are extracted and rendered onto the input canvas using the identical randomized rendering strategy described above. 
Because ImageNet inherently contains multiple diverse target images for each class, a single rendered label canvas can be paired with various target images from the same category. 
This one-to-many pairing strategy naturally encourages the model to capture and generate intra-class diversity.
\paragraph{Text-in-Image Editing.}
This category serves as the cornerstone for our model's instruction-following capabilities, systematically unifying semantic image manipulations and structural condition-to-image tasks into a single learning objective. 

To cover a comprehensive spectrum of operational intents, we aggregate data from four major sources: GPT-Image-Edit~\cite{wang2025gptimageedit15mmillionscalegptgeneratedimage}, Pico-Banana~\cite{qian2025picobanana400klargescaledatasettextguided}, UnicEdit~\cite{ye2025unicedit}, and PixWizard~\cite{lin2024pixwizard}. 
Instead of enumerating all fine-grained sub-tasks, we conceptually group the diverse editing capabilities into several core dimensions:
\begin{itemize}
    \item \textbf{Semantic Operations:} Including subject addition, removal, replacement, and object swapping.
    \item \textbf{Attribute \& Environment Modifications:} Covering local property changes (color, material, age/gender, facial expressions) and global atmospheric adjustments (lighting, weather conditions, background swapping).
    \item \textbf{Artistic \& Style Transfer:} Ranging from fundamental stylization to highly specific domain translations (e.g., 2D anime, Pixar-like 3D, sketch, line-art, and Western comic styles).
    \item \textbf{Complex Spatial Reasoning:} Encompassing multi-object coordination, counting changes, outpainting, and pose adjustments.
\end{itemize}

While traditional paradigms treat structural condition-to-image generation (e.g., Depth-to-image and segmentation-to-image) or restoration tasks (inpainting, face/nature restoration) as distinct architectural branches, we argue that they naturally fall under the umbrella of ``image editing''.
Specifically, a structural condition map (such as a Canny edge map) is inherently a source image. By treating these spatial conditions as the starting canvas, we seamlessly integrate 315K high-quality pairs from PixWizard into our unified training pipeline.

To fully align with the purely vision-centric \modelname architecture, all operational intents must be explicitly embedded as visual text prompts.
We transform the conventional heterogeneous triplet $\langle \text{text instruction, source image, target image} \rangle$ into a strictly visual $\langle \text{input image, target image} \rangle$ pair. 
This is achieved by directly rendering the textual instruction onto the source image (or condition map) being edited.
Consistent with our fundamental generation strategy, we apply extensive randomized augmentations during this rendering process—randomly sampling the font style, text size, font color, and spatial placement on the canvas.
This guarantees that the model learns to robustly perceive and parse visual text commands in varied scenarios rather than relying on fixed typographical shortcuts.

To maintain a high-quality optimization landscape for flow matching, we implement a strict filtering mechanism on the combined 2.5M raw semantic editing pairs, discarding complex or inconsistent pairs.
Specifically, we filter out examples exhibiting: (1) highly ambiguous or overly convoluted textual instructions that lack explicit visual targets; (2) source images with severe generative artifacts or extreme aspect ratios; and (3) poor visual-semantic alignment, where the target image fails to faithfully reflect the specific operational intent dictated by the text. 
After filtering, we retain approximately 1.6M highly curated semantic editing pairs, yielding a total of 1.9M unified pairs when combined with PixWizard.

\paragraph{Text Bounding Box Editing.}
To endow the model with precise spatial control, we construct a high-quality subset where text and bounding boxes jointly dictate the generation. 
Rather than utilizing existing imperfect pairs, we sample 45K high-resolution, artifact-free, and aesthetically pleasing images from GPT-Image-Edit. These images serve solely as the initial unedited source images.

\textbf{Two-Stage Vision-Guided Synthesis:} 
The generation of valid image pairs follows a customized pipeline leveraging the combined priors of Qwen Image Edit~\cite{wu2025qwenimagetechnicalreport} and Qwen3-VL~\cite{Qwen2.5-VL}.
First, we randomly define a set of target objects to be added. We then prompt Qwen Image Edit to seamlessly insert the specified object into the source canvas, generating the final target image. 
Subsequently, we employ Qwen3-VL as a visual grounding agent to precisely localize the newly inserted object within the target image.
Based on these spatial coordinates, we draw a prominent bounding box directly onto the original source images and render text label of the object adjacent to the box. 
In this purely visual prompt configuration, the textual label explicitly dictates the semantic category, while the bounding box imposes strict geometric constraints defining the scale and location.

\textbf{Automated Logit-Based Filtering:} 
To ensure the highest data quality and eliminate the need for laborious manual inspection, we design an automated auditing mechanism powered by Qwen3-VL 7B.
The auditor evaluates the generated pairs across three critical dimensions: instruction faithfulness, local visual consistency, and global background preservation. 
Crucially, rather than relying on simple binary outputs, the auditor computes a continuous confidence score based on output logits:
\begin{equation}
    Score = \frac{P(\text{Yes}) - P(\text{No})}{P(\text{No})}
\end{equation}
This formulation allows us to establish a strict, customizable threshold for data retention. 
When an edit is deemed unsuccessful (i.e., falls below the threshold), the auditor activates a Refinement Protocol. 
It identifies specific synthesis issues (e.g., color mismatch or texture corruption) and outputs a prompt starting with ROP (Refinement Output Prompt) to provide feedback for the generation pipeline. 
By applying this rigorous auditing standard, we filter out sub-optimal generations and ultimately retain 24K high-fidelity pairs that exhibit perfect spatial-visual alignment.
\begin{figure*}[htbp]
    \centering
    \includegraphics[width=\textwidth]{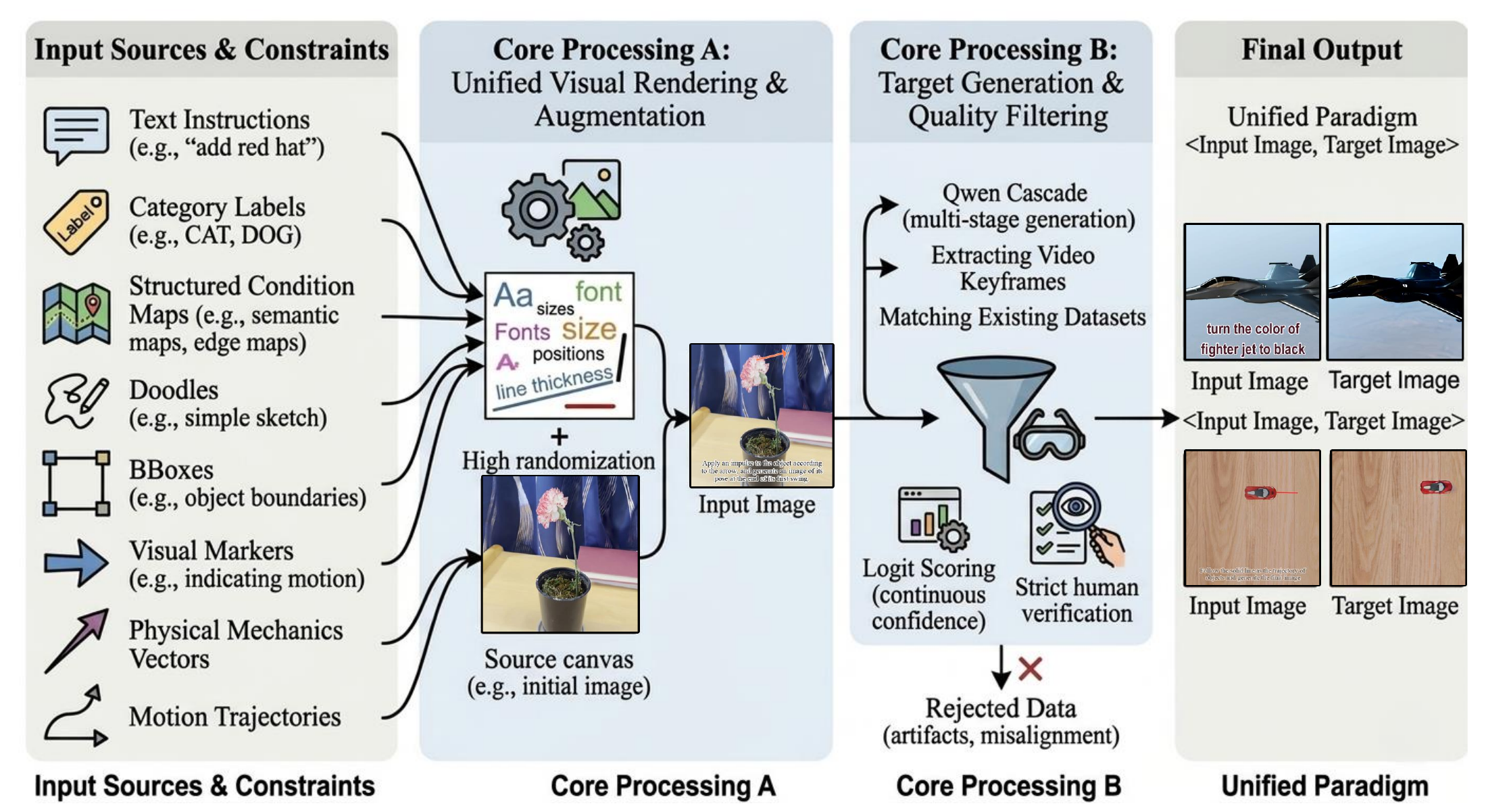}
    \caption{\textbf{General data construct process.}}
    \label{fig:data_construct_process}
\end{figure*}
\paragraph{Visual Marker Editing.}
Visual markers, such as arrows, offer a highly intuitive and efficient interface for human users. 
They enable precise spatial referencing without requiring explicit object names, which is particularly advantageous when the target object is ambiguous or difficult to describe. 
To construct this subset, we curate high-quality image pairs from UltraEdit~\cite{zhao2024ultraedit} and OmniEdit~\cite{wei2024omniedit}, focusing specifically on operation-centric sub-categories including \textit{removal, replacement, object swap, attribute modification}, and \textit{color/local changes}. 

To automate the visual marker annotation, we first parse the original textual instructions to extract the specific target subject. 
We then employ Qwen3-VL to precisely ground this subject within the input image. Based on the spatial coordinates, a script dynamically renders an arrow on the source canvas.
To prevent the model from overfitting to specific marker styles, the arrow's color, size, and starting position are completely randomized, with the strict constraint that its tip must point directly at the target object.
Finally, the explicit object name in the instruction is systematically replaced with a generic visual reference (e.g., ``the object pointed by the arrow'').
To guarantee data quality, we subject these synthesized pairs to the exact same automated logit-based auditing mechanism detailed in the Text \& BBox Guided Editing section.
This rigorous filtering process ultimately yields 250K highly reliable pairs, forcing the model to explicitly comprehend semantics and spatial relationships through visual cues alone.
\vspace{-2mm}
\paragraph{Doodles Editing.}
Doodles provide an intuitive interface for users to explicitly specify shape priors and spatial layouts. 
To construct this subset, we collect 5K high-quality web images to serve as unedited base canvases and predefine ten diverse object categories. We then employ a two-stage synthesis pipeline powered by Qwen Image Edit~\cite{wu2025qwenimagetechnicalreport}. 
In the first stage, the model is prompted to insert 1 or 2 simple, hand-drawn style doodles from the predefined categories into the base canvas, forming the \textit{input image}. In the second stage, Qwen Image Edit transforms these abstract doodles into photo-realistic objects, yielding the corresponding \textit{target image}.

Due to the inherent generative instability of multi-stage image-to-image translation, a significant portion of the initial outputs suffer from structural errors. 
Therefore, we conduct a rigorous manual inspection. The retention criteria strictly dictate that: (1) there must be an absolute absence of generative artifacts; (2) the doodles must remain structurally simple and abstract, rather than prematurely resembling real objects; (3) the synthesized photo-realistic objects must perfectly align with the shape priors defined by the doodles; and (4) the non-edited regions must maintain perfect pixel-level consistency with the input image. 
This meticulous manual curation filters out the majority of failure cases, ultimately yielding 1K high-fidelity pairs.
\paragraph{Force Understanding.}
To endow \modelname with physics-aware reasoning, we utilize the Force Prompting dataset~\cite{gillman2025forcepromptingvideogeneration}, which encompasses two distinct physical paradigms: \textit{point forces} and \textit{global forces}. 
Specifically, the point force subset captures the linear kinematics of spherical objects and the harmonic oscillations of plants, while the global force subset simulates aerodynamic effects, such as wind acting on flags. 
To transform these continuous video dynamics into static image-to-image reasoning pairs, we implement a systematic, physics-driven keyframe extraction strategy. 
We designate the first frame of each video as the initial state canvas. 
For the target outcome, rather than arbitrary sampling, we employ specific analytical strategies (e.g., optical flow analysis) to determine the temporal point of kinematic convergence---a statistically determined frame where the dynamic systems reliably reach a terminal state, steady equilibrium, or maximum physical displacement. 

Finally, we explicitly visualize the underlying physical parameters.
The original dataset provides precise annotations: the exact spatial coordinate of the applied force, the force angle, and a normalized magnitude scalar $m \in [0, 1]$. Utilizing an automated script, we render explicit force arrows directly onto the initial frame. For point forces, the arrow originates precisely at the application coordinate. 
For global forces, it is rendered as a global environmental indicator. 
The arrow's geometric length is strictly proportional to the magnitude $m$, and its orientation aligns with the physical angle. 
This fully annotated initial frame forms our \textit{input image}, explicitly challenging the model to predict the corresponding steady-state \textit{target image} strictly governed by the visualized force dynamics.
\paragraph{Trajectory Understanding.}
To further explicitly model physics-aware motion priors, we construct a highly curated trajectory understanding subset. 
We first manually render a collection of high-fidelity dynamic videos featuring car and ball movements using the Blender 3D engine, encompassing both linear and complex curved kinematics. 

To formulate the static image-to-image pairs, we extract the initial and terminal frames of each video. 
Our manual annotation protocol requires human annotators to draw a solid, continuous line directly onto the initial frame, accurately tracing the exact future geometric path of the moving object.
This meticulously annotated frame serves as the \textit{input image}, while the original terminal frame (showing the object at its final destination) acts as the corresponding \textit{target image}. 
Crucially, to ensure the model strictly learns the spatial geometric path rather than overfitting to specific visual artifacts, the thickness and color of the drawn trajectory lines are entirely randomized during the annotation process.
This rigorous manual pipeline ultimately yields 1.5K highly precise pairs for motion trajectory generation.

\subsection{Visual Text Rendering Pipeline}
\label{subsec:rendering_pipeline}

To synthesize high-fidelity and diverse visual instruction data, we propose a robust, automated text-rendering engine.
The pipeline is designed to dynamically adapt to various text lengths and underlying background constraints while strictly preserving legibility and geometric alignment. The rendering process comprises five pivotal stages:

\vspace{0.5em}
\noindent\textbf{1. Robust Font Selection and Glyph Validation.}
To ensure the generative robustness of the text rendering, we implement a dynamic font-picking mechanism. 
Given an input text sequence, the engine first validates character support by parsing the TrueType font's \texttt{cmap} tables. 
To prevent the rendering of corrupted or ``missing glyph'' boxes (often caused by incomplete font files), we introduce an empirical glyph-area validation threshold. 
Let $A_c$ be the bounding box area of a rendered character $c$ and $s$ be the font size. A font is deemed robust and selected only if the average active pixel ratio $\frac{1}{N} \sum_{c \in T} \frac{A_c}{s^2}$ exceeds a predefined minimum threshold, guaranteeing high-quality typographic representation across millions of synthesized pairs.

\vspace{0.5em}
\noindent\textbf{2. Semantic-Aware Tokenization.}
Handling multi-lingual instructions requires precise line-breaking strategies. We utilize a custom tokenization algorithm tailored for visual layouts. 
Characters are isolated as individual tokens to allow flexible word wrapping, whereas Latin alphanumeric sequences and symbols are grouped as cohesive whole-word tokens. 
This strategy prevents improper truncations of Western words at the end of a line, strictly preserving the semantic readability of the visual prompt.

\vspace{0.5em}
\noindent\textbf{3. Adaptive Bounding-Box Layout Algorithm.}
To automatically determine the optimal typographic layout within a constrained visual canvas, we model the layout generation as a constrained optimization problem. 
Given a target bounding box with dimensions $W \times H$, our goal is to find the maximum font size $s_{max}$ that accommodates the tokenized sequence $T$ without overflow. 
We solve this efficiently in $O(\log N)$ time using a binary search algorithm over the font size space $[s_{min}, s_{max}]$. 
For instances with extensive token counts, the engine defaults to utilizing the maximum available canvas margin.
For shorter instructions, we introduce spatial diversity by randomizing the location and dimensions of the localized bounding boxes, thereby forcing the generative model to understand text instructions across arbitrary spatial distributions.
\begin{algorithm}[tb]
\caption{Adaptive Bounding-Box Layout via Binary Search}
\label{alg:layout_search}
\textbf{Input:} Token sequence $T$, Target bounding box dimensions $W \times H$, Font size search space $[S_{min}, S_{max}]$ \\
\textbf{Output:} Optimal font size $S^*$, Layout configuration $L^*$ (or \text{Null} if infeasible)
\begin{algorithmic}[1] 
\State $low \gets S_{min}$, $high \gets S_{max}$
\State $S^* \gets \text{Null}$, $L^* \gets \text{Null}$
\While{$low \le high$}
    \State $mid \gets \lfloor (low + high) / 2 \rfloor$
    \State $L_{tmp} \gets \text{TryWordWrap}(T, W, mid)$ \Comment{Greedy wrap within width $W$}
    
    \If{$L_{tmp} \neq \text{Null}$} \Comment{Valid width: No single token exceeds $W$}
        \State $H_{tmp} \gets \text{CalculateTotalHeight}(L_{tmp}, mid)$
        \If{$H_{tmp} \le H$} \Comment{Valid height: Text fits entirely within $H$}
            \State $S^* \gets mid$ \Comment{Update optimal states}
            \State $L^* \gets L_{tmp}$
            \State $low \gets mid + 1$ \Comment{Attempt to maximize readability (larger font)}
        \Else
            \State $high \gets mid - 1$ \Comment{Height overflow, require smaller font}
        \EndIf
    \Else
        \State $high \gets mid - 1$ \Comment{Width overflow, require smaller font}
    \EndIf
\EndWhile
\State \Return $S^*, L^*$
\end{algorithmic}
\end{algorithm}
\vspace{0.5em}

\noindent\textbf{4. Context-Aware Stylization and Alpha Compositing.}
To guarantee text legibility regardless of the underlying visual content, we integrate a context-aware color contrast mechanism. Before rendering, the engine calculates the perceptual luminance $L$ of the underlying image region bounded by the text block:
\begin{equation}
    L = 0.299 \mu_R + 0.587 \mu_G + 0.114 \mu_B
    \label{eq:luminance}
\end{equation}
where $\mu_R, \mu_G, \mu_B$ denote the mean channel intensities of the cropped background. 
If the local background is heavily illuminated ($L > 128$), the engine applies dark text fill coupled with a thick white stroke; conversely, it utilizes bright text with a dark stroke for low-luminance regions.
The stroke width is dynamically scaled based on the calculated line height. Finally, the text is rendered onto a dedicated transparent RGBA layer and seamlessly merged with the base canvas using alpha compositing, eliminating visual artifacts along the font anti-aliasing edges.

\vspace{0.5em}
\noindent\textbf{5. Analysis of the Layout Algorithm.} 
The proposed adaptive layout strategy (Algorithm \ref{alg:layout_search}) provides several critical advantages for large-scale data synthesis:

\begin{itemize}
    \item \textbf{Computational Efficiency:} Traditional text rendering engines often rely on a linear step-down approach (iteratively decreasing font size until the text fits), yielding a time complexity of $O(S_{max} - S_{min})$. By formulating the layout process as a binary search optimization, we reduce the complexity to $O(\log(|S_{max} - S_{min}|))$. This logarithmic efficiency is paramount when dynamically rendering over 5 million high-resolution image pairs, significantly accelerating the data generation pipeline.
    
    \item \textbf{Robust Fallback Mechanism:} For extreme edge cases—such as exceptionally long instructions or single words that exceed the randomized bounding box width $W$ even at $S_{min}$—the algorithm seamlessly triggers a global fallback. Instead of discarding these valuable data points, the engine automatically re-initializes the target dimensions $W \times H$ to the maximum safe canvas margins. This hierarchical container strategy guarantees a near 100\% layout success rate, preventing long-tailed complex instructions from being systematically filtered out.
    
    \item \textbf{Spatial Variance as Implicit Augmentation:} By stochastically sampling the initial dimensions $(W, H)$ and the starting anchor coordinates $(X, Y)$ rather than always utilizing the full canvas, we introduce vast spatial diversity. This design forces the downstream generative model to learn robust spatial grounding and positional alignment, ensuring that the model adheres to precise geometric constraints rather than simply memorizing centered, full-screen text overlays.
\end{itemize}

\subsection{Automated Quality Control and Filtering Pipeline}
\label{sec:quality_control}

This section provides a granular breakdown of the automated filtering pipelines, the statistical properties of the curated data, and the rigorous evaluation protocols employed in our benchmark.

Given that our raw image pairs are sourced from diverse public datasets (yielding highly variable initial quality), we implement a rigorous, multi-stage filtering pipeline. Following the large-scale visual text rendering detailed in Section \ref{subsec:rendering_pipeline}, the generated images undergo a comprehensive inspection to guarantee visual fidelity, text legibility, and data diversity.

\vspace{0.5em}
\noindent\textbf{1. OCR-based Legibility Verification.}
To ensure the synthesized text is completely legible and free from truncation or rendering artifacts (e.g., overlapping bounding boxes or corrupted glyphs), we deploy an Optical Character Recognition (OCR) engine as the first filter. Let $T_{src}$ denote the original instruction text and $T_{ocr}$ denote the text extracted from the rendered canvas $I_v$. We compute the Character Error Rate (CER) and filter out pairs where the error exceeds a stringent threshold $\tau_{ocr}$:
\begin{equation}
    \text{CER}(T_{src}, T_{ocr}) = \frac{S + D + I}{N} \le \tau_{ocr}
\end{equation}
where $S, D,$ and $I$ are the number of substitutions, deletions, and insertions, respectively, and $N$ is the total number of characters in $T_{src}$. Images failing this check are discarded to prevent the model from learning corrupted visual instructions.
\begin{figure*}[!t]
    \centering
    \includegraphics[width=\textwidth]{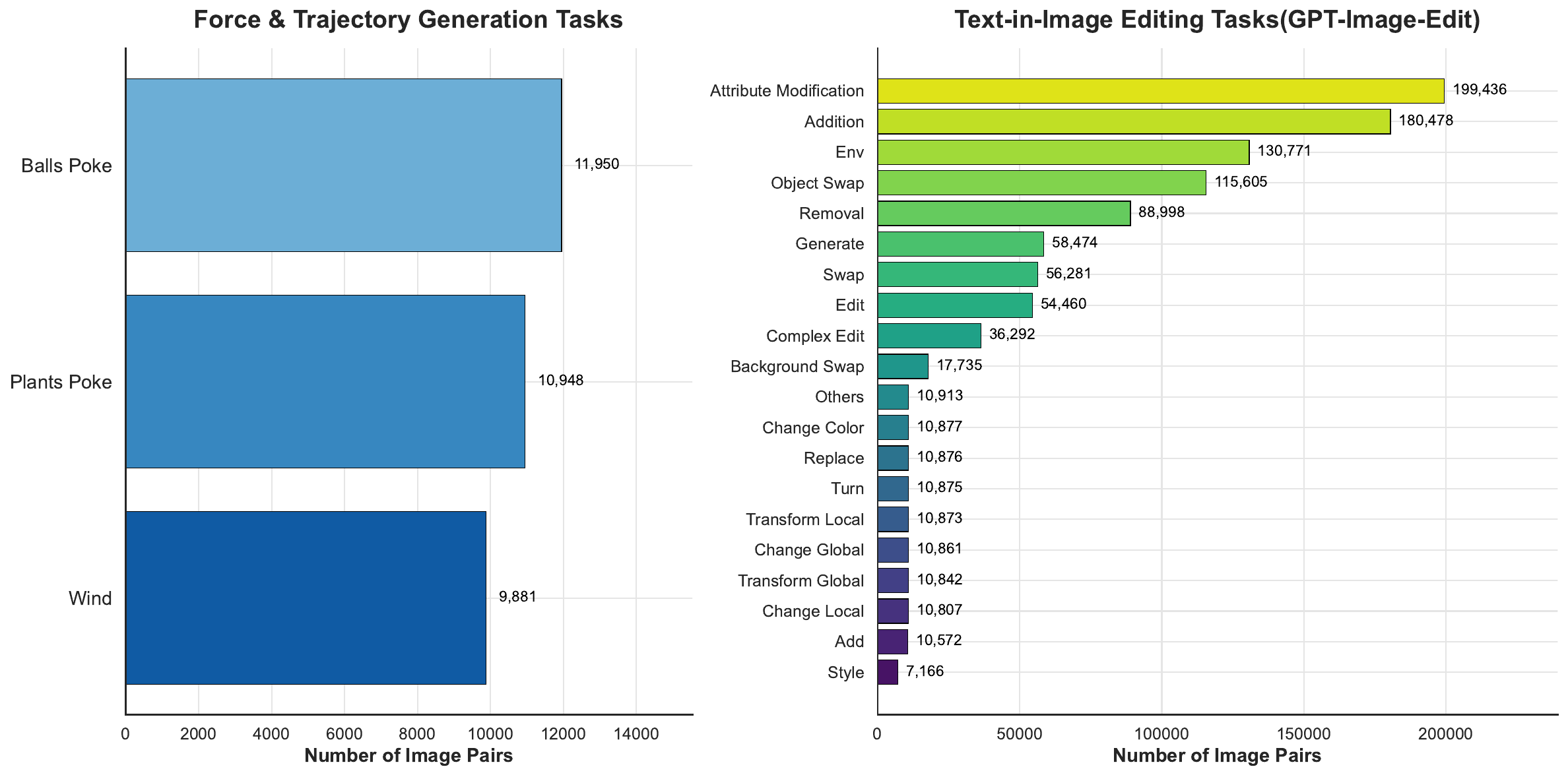}
    \caption{\textbf{Detailed instance distribution of \datasetname.} \textbf{Left:} The Force \& Trajectory Generation subset, highlighting strictly curated physics-aware categories (e.g., wind, object poking) designed to impart dynamic kinematic priors. \textbf{Right:} The Text-in-Image Editing subset (derived from GPT-Image-Edit), demonstrating a natural long-tailed distribution of semantic operations. Ranging from high-frequency attribute modifications to specialized style transfers, this structural diversity ensures robust generalization across both complex stylistic changes and physical constraints.}
    \label{fig:dataset_distribution1}
\end{figure*}

\noindent\textbf{2. Task-Specific VLM Quality Inspection.}
Images that pass the OCR check are subsequently evaluated by an advanced Multimodal Large Language Model (MLLM, e.g., Qwen3-VL). To handle the diverse nature of our generative tasks, we design task-specific prompts. The VLM acts as a judge, outputting a boolean decision based on customized criteria:
\begin{itemize}
    \item \textit{Fundamental Generation:} ``Does the main subject in the image perfectly align with the embedded text prompt: [PROMPT]?''
    \item \textit{Spatial Constraints (BBoxes/Markers):} ``Is the object precisely located within the red bounding box/indicated by the visual arrow?''
    \item \textit{Physics-Aware Operations:} ``Does the motion blur or trajectory accurately reflect the directional force specified by the vector arrows?''
\end{itemize}
Only pairs that receive a positive confirmation across both semantic alignment and visual realism are retained.

\noindent\textbf{3. Diversity-Oriented Deduplication.}
To maximize the informational entropy of the dataset and prevent mode collapse during training, we apply a diversity-oriented filtering mechanism. We extract CLIP image embeddings $E_{clip}(I)$ for all candidates within a specific sub-task. A candidate $I_i$ is retained only if its cosine similarity with all previously accepted images $I_j$ in the active pool $\mathcal{P}$ remains below a diversity threshold $\tau_{div}$:
\begin{equation}
    \max_{I_j \in \mathcal{P}} \left( \frac{E_{clip}(I_i) \cdot E_{clip}(I_j)}{\|E_{clip}(I_i)\| \|E_{clip}(I_j)\|} \right) < \tau_{div}
\end{equation}
This strategy effectively prunes redundant concepts, ensuring a highly diverse data distribution.

\subsection{Dataset Composition and Detailed Statistics}
\label{subsec:dataset_composition}

Through the aforementioned rendering and rigorous filtering pipeline, we curated a final dataset of approximately 5M high-quality image pairs. 
Table \ref{tab:dataset_composition} details the macro-level composition, primary sources, and final retention volumes for each major task category. 
To further illustrate our stringent quality control, Table \ref{tab:filtering_rates} provides a breakdown of the filtering survival rates across selected complex generative categories.

\begin{figure*}[htbp]
    \centering
    \includegraphics[width=\textwidth]{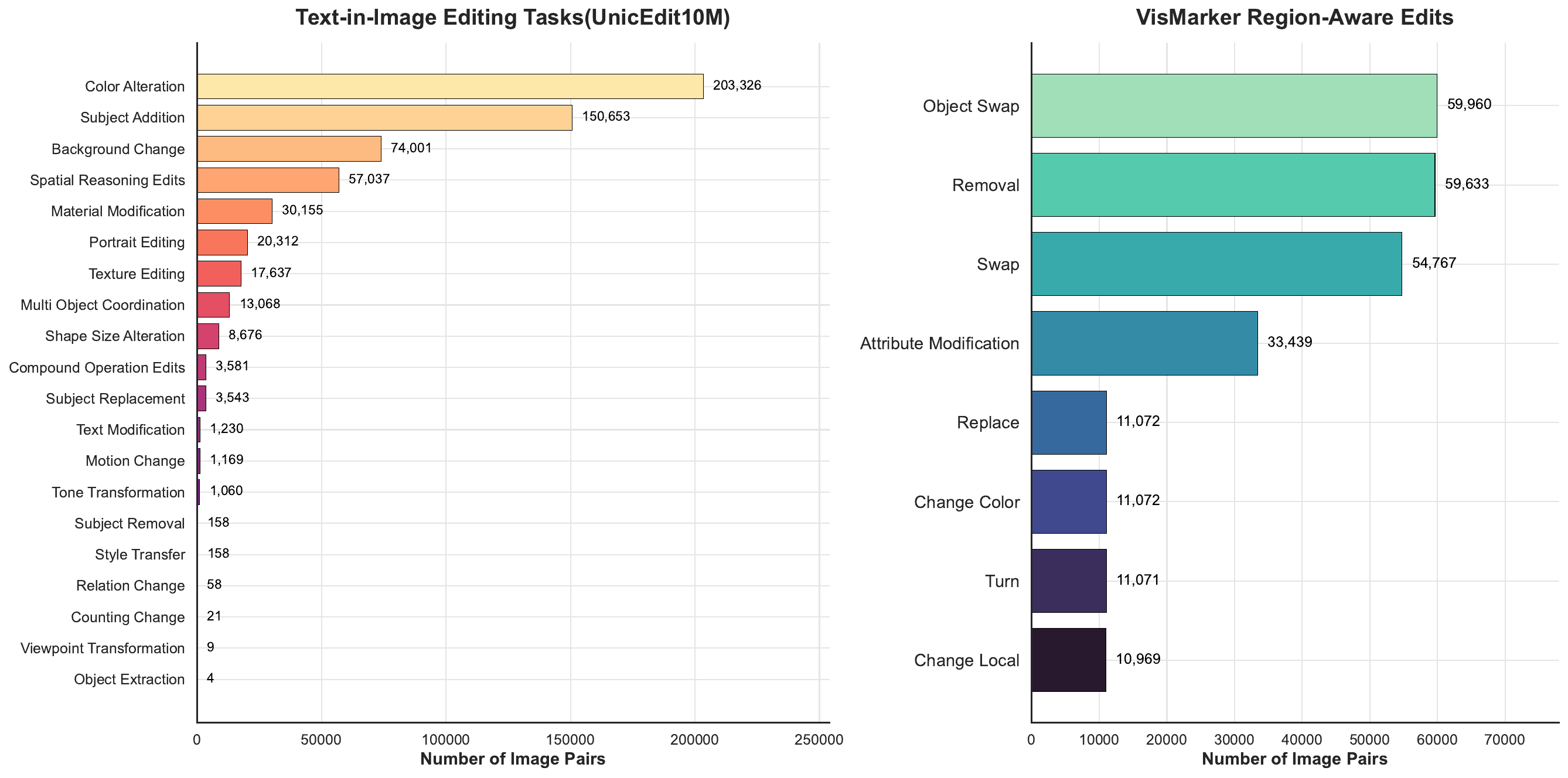}
    \caption{\textbf{Semantic diversity and region-aware distributions of \datasetname.} \textbf{Left:} The UnicEdit10M Diverse Edits subset, showcasing a profound long-tailed distribution. It spans high-frequency semantic operations like Color Alteration and Subject Addition to rare edge cases such as Object Extraction. \textbf{Right:} The VisMarker Region-Aware Edits subset, featuring robust, high-volume spatial operations like Object Swap and Removal. This combination ensures the model learns broad semantic reasoning while maintaining strict adherence to localized visual markers.}
    \label{fig:dataset_distribution3}
\end{figure*}

\begin{figure*}[htbp]
    \centering
    \includegraphics[width=\textwidth]{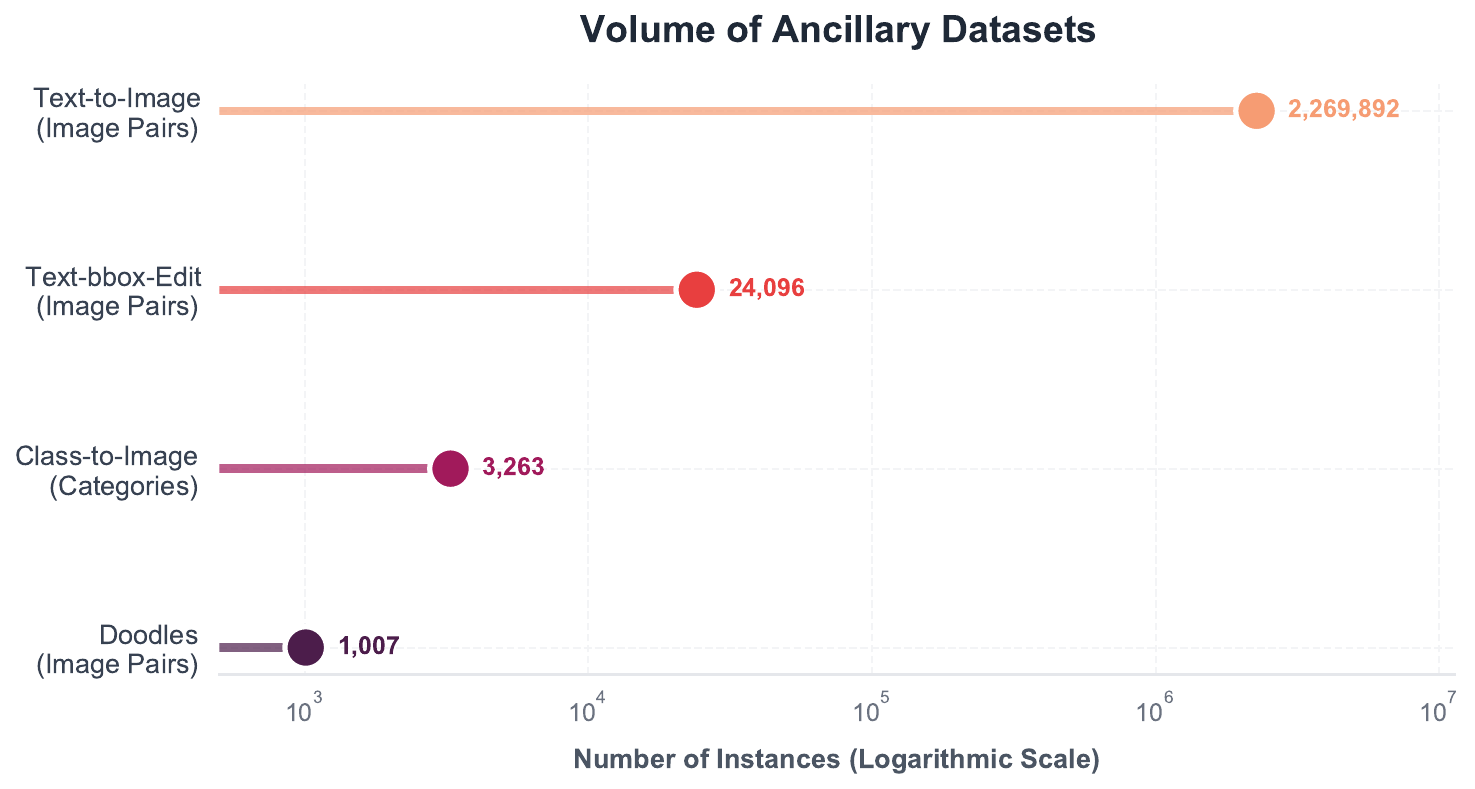}
    \caption{\textbf{Volume of ancillary generation and structural datasets.} This logarithmic lollipop chart illustrates the extreme variance in scale and specialization across our supplementary sources. It highlights the massive foundation of over 2.26 million general Text-to-Image pairs, balanced by broad categorical coverage (ImageNet21K) and highly specialized, precise structural tasks such as Text Bbox Edit and Doodles. This multi-scale composition ensures the model achieves both robust open-domain generative capabilities and fine-grained execution of explicit visual constraints.}
    \label{fig:dataset_distribution4}
\end{figure*}

Beyond macroscopic volumes, analyzing the intra-category distributions is crucial for understanding the structural diversity of \datasetname. As illustrated in our distribution figures (refer to Figure \ref{fig:dataset_distribution1},~\ref{fig:dataset_distribution2},~\ref{fig:dataset_distribution3},~\ref{fig:dataset_distribution4}), the curated dataset exhibits a multi-granularity structure tailored to impart distinct generative priors to the model:

\begin{figure*}[!t]
    \centering
    
    \includegraphics[width=\textwidth]{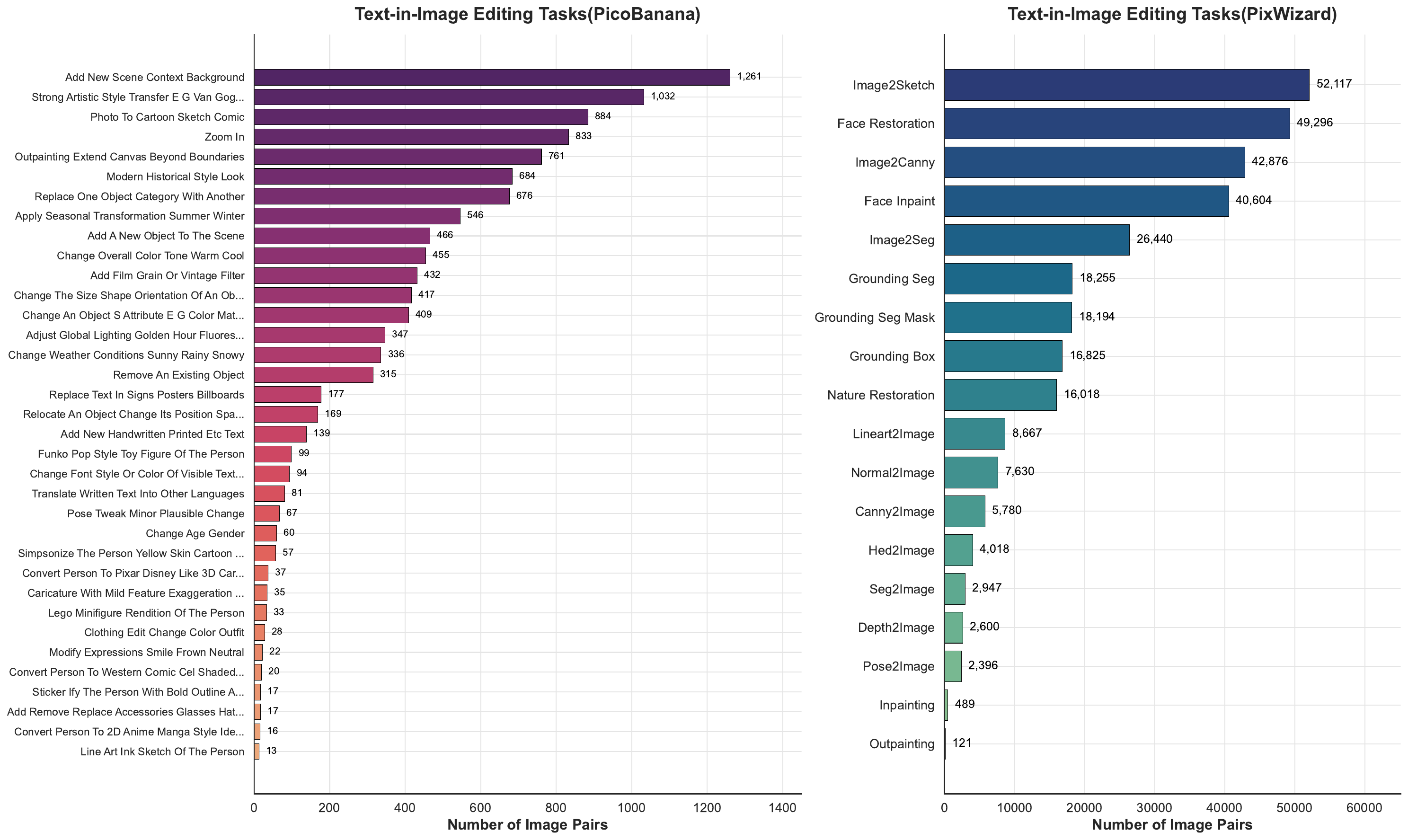}
    \caption{\textbf{Fine-grained structure and stylized distributions of \datasetname.} \textbf{Left:} The Structured Editing subset (derived from PixWizard), highlighting dense spatial translations such as Image-to-Sketch and Face Restoration. This subset trains the model to strictly adhere to geometric and structural conditions. \textbf{Right:} The Text-in-Image Editing subset (derived from PicoBanana), detailing 35 highly specialized, long-tailed aesthetic operations. This extreme categorical diversity ensures the model's proficiency in handling nuanced, localized, and composite textual instructions.}
    \label{fig:dataset_distribution2}
\end{figure*}

\vspace{0.5em}
\noindent\textbf{1. Broad Semantic and Stylistic Coverage (Long-Tailed Nature).} 
Our text-in-image editing subsets—derived heavily from UnicEdit, GPT-Image-Edit, and PicoBanana—encompass a massive spectrum of user intents. UnicEdit and GPT-Image-Edit contribute the bulk of the volume, dominated by high-frequency operations such as \textit{Color Alteration} ($\sim$203K), \textit{Attribute Modification} ($\sim$199K), and \textit{Subject Addition} ($\sim$150K). Interestingly, the data naturally exhibits a long-tailed distribution (e.g., rare tasks like \textit{Object Extraction} or \textit{Counting Change} in UnicEdit contain fewer than 100 pairs). Rather than artificially flattening this curve, we intentionally preserve this long-tailed property to reflect real-world human editing priors. Furthermore, PicoBanana injects extreme stylistic diversity, contributing 35 highly specialized, fine-grained categories (e.g., \textit{Simpsonize}, \textit{Vintage Filter}, \textit{Outpainting}), ensuring the model's robustness against complex, composite textual instructions.

\vspace{0.5em}
\noindent\textbf{2. Spatial Reasoning and Region-Aware Constraints.} 
While text instructions govern semantic changes, visual and geometric inputs dictate spatial precision. Our structured editing subsets (PixWizard and VisMarker) serve this exact purpose. The VisMarker subset provides highly balanced, region-aware supervision across 8 core categories (e.g., \textit{Object Swap}, \textit{Removal}, ranging consistently between 33K and 60K pairs), forcing the model to strictly respect local visual markers rather than applying global stylistic shifts. Concurrently, the PixWizard subset injects dense structural conditions, containing robust distributions of \textit{Face Restoration} ($\sim$49K) and \textit{Image-to-Sketch} ($\sim$52K), which train the model to understand dense spatial mappings like bounding boxes and segmentation masks unified within the input canvas.

\vspace{0.5em}
\noindent\textbf{3. Physics-Aware and Kinematic Dynamics.} 
A uniquely challenging component of \textbf{\datasetname} is the Force \& Trajectory generation subset. While smaller in scale compared to semantic edits (comprising specifically curated classes like \textit{balls\_poke} at $\sim$11K and \textit{wind} at $\sim$9K), this subset is of exceptionally high fidelity. It forces the image-to-image paradigm to step beyond static pixel manipulation and understand dynamic kinematic priors, translating explicit visual force vectors (arrows and magnitudes) into physically plausible consequences like motion blur, structural deformation, and trajectory extrapolation.

\vspace{0.5em}
In summary, the statistical distribution of \datasetname is purposefully engineered. The massive text-in-image editing pairs provide a robust semantic foundation, the structured marker datasets enforce spatial discipline, and the curated physics subset unlocks novel dynamic capabilities, collectively empowering a single model to master multi-modal, instruction-driven image generation.
\begin{table}[ht]
\centering
\vspace{-3mm}
\caption{Comprehensive breakdown of the \datasetname composition. The dataset aggregates various public sources, strictly refined through our multi-stage OCR and VLM filtering pipeline.}
\label{tab:dataset_composition}
\resizebox{\linewidth}{!}{%
\begin{tabular}{llrr}
\toprule
\textbf{Task Category} & \textbf{Primary Source(s)} & \textbf{Initial Volume} & \textbf{Curated Pairs} \\ 
\midrule
Fundamental Gen. (Text) & Text-to-Image-2M~\cite{zk_2024} & 2,269,892 & 2,269,892 \\
Fundamental Gen. (Class)& ImageNet Subset~\cite{Russakovsky2014ImageNetLS}& 2,000,000 & 860,000 \\
Text-in-Image Editing   & GPT-Image-Edit, Pico-Banana, UnicEdit & 2,100,000 & 1,601,000 \\
Structured Editing      & PixWizard~\cite{lin2024pixwizard} & 350,000 & 315,000 \\
Bounding Box Editing    & GPT-Image-Edit (Filtered) & 45,000 & 24096 \\
Visual Marker Editing   & Qwen3-VL Synthesized & 300,000 & 250,000 \\
Doodles Editing         & Web Crawled Images & 5,000 & 1,007\\
Trajectory Understanding  & Blender Renders  & 1,600 & 1,513 \\ 
Force Understanding     &  Force Prompting~\cite{gillman2025forcepromptingvideogeneration} & 36,000 & 25,820 \\
\midrule
\textbf{Total}          & - & \textbf{$\sim$7.12M} & \textbf{$\sim$5.36M} \\ 
\bottomrule
\end{tabular}%
}
\end{table}

\begin{table}[ht]
\centering
\caption{Filtering survival rates across selected complex generative tasks, demonstrating the rigorousness of our automated OCR and VLM inspection.}
\label{tab:filtering_rates}
\resizebox{\linewidth}{!}{%
\begin{tabular}{lccc}
\toprule
\textbf{Category} & \textbf{Primary Rejection Reason} & \textbf{VLM/OCR Filtering} & \textbf{Retention Rate} \\ 
\midrule
Text-in-Image Editing & Semantic Inconsistency & VLM Semantic Check & 78.2\% \\
Bounding Box Editing  & Geometric Misalignment & VLM Spatial Check & 53.3\% \\
Doodles Editing       & Generative Instability & VLM Realism Check & 20.0\% \\
Visual Marker Editing    & Generative Instability & VLM Spatial Check & 83.3\% \\
\bottomrule
\end{tabular}%
}
\vspace{-5mm}
\end{table}

\section{Limitations and future work}
\label{AppendE}
While our model introduces a promising unified paradigm for visual instruction following, we acknowledge several limitations in the current framework. First, although the model demonstrates strong performance on our benchmark, its generalization capabilities in highly complex, unconstrained scenarios remain somewhat limited. 
This is primarily bounded by our current model capacity (1.2B parameters) and the scale of the training dataset. 
Second, due to computational constraints during training, the output generation is currently restricted to a fixed spatial resolution of $256 \times 256$ pixels, which may not fully satisfy the demands of high-fidelity creative workflows. 
Finally, our approach is currently optimized for single-turn instruction execution, and its potential for continuous, multi-turn interactive editing has yet to be fully explored. 
In future work, we aim to scale up both the model parameters and the training data to handle increasingly complex scenarios, further optimize our framework to support high-resolution generation, and extend our visual-centric paradigm to facilitate seamless multi-turn visual editing scenarios.
\section{More evaluation details}
\label{AppendF}
\subsection{VLM evaluation}
\label{vlm_eval}
\begin{figure*}[!t]
    \centering
    \includegraphics[width=\textwidth]{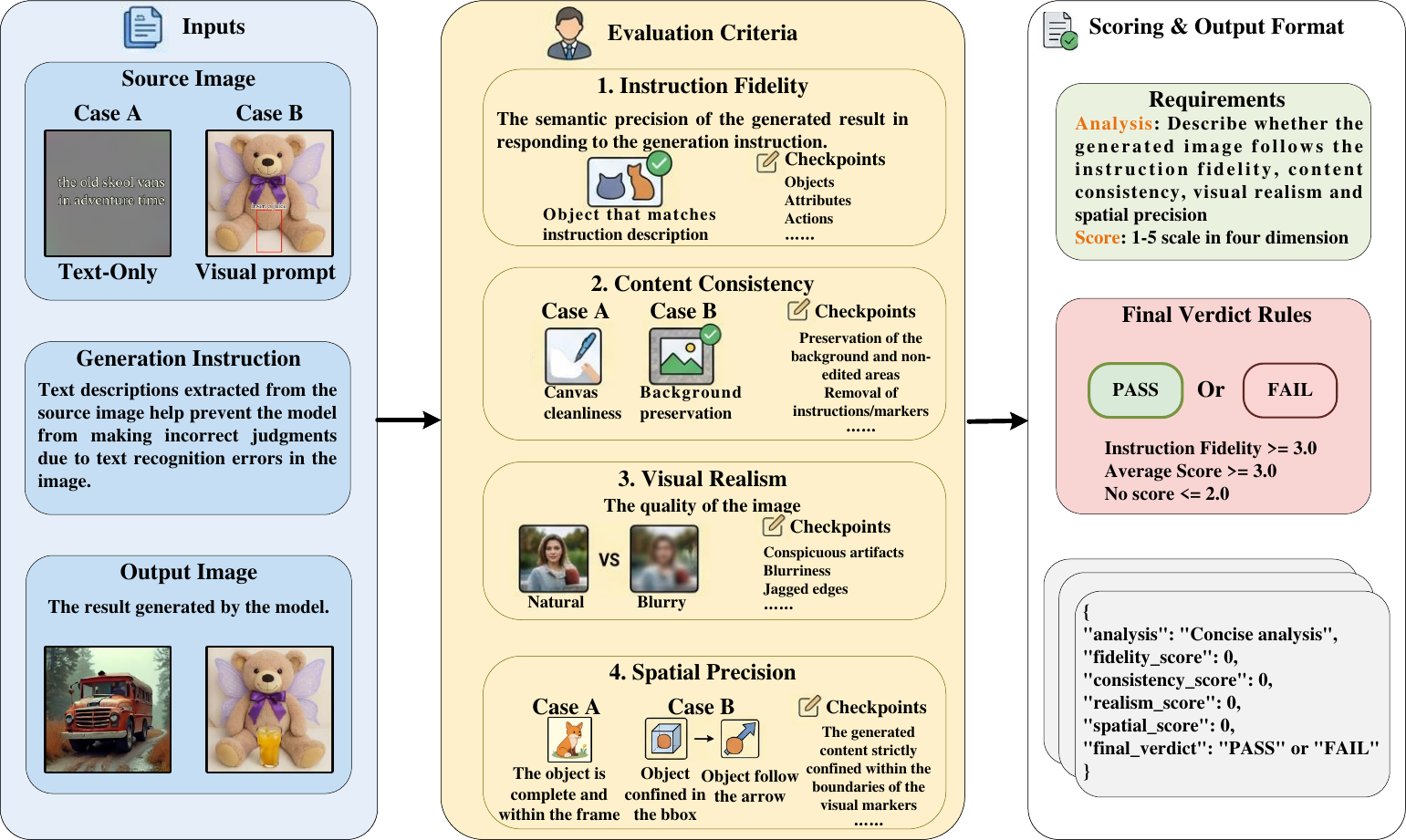}
    \caption{Evaluation process.}
    \label{fig:eval_process}
\end{figure*}

To ensure a comprehensive, objective, and reproducible assessment of visual instruction following, we design a systematic evaluation pipeline driven by Vision-Large Language Models (VLMs). As illustrated in Figure~\ref{fig:eval_process}, our evaluation process takes three primary inputs: the source image (categorized into Case A for text-only canvases and Case B for annotated real-world images), the generated output image, and a plain text generation instruction. This text instruction is explicitly extracted from the source image to prevent the VLM judge from making incorrect judgments due to inherent Optical Character Recognition (OCR) errors in the image.

The VLM evaluates the generated images across four distinct criteria on a 1-5 scale:
\begin{itemize}
    \item \textbf{Instruction Fidelity:} Measures the semantic precision of the generated result (e.g., matching objects, attributes, and actions) in responding to the core generation instruction.
    \item \textbf{Content Consistency:} For generation tasks (Case A), this evaluates canvas cleanliness. For editing tasks (Case B), it strictly checks for the preservation of unedited background regions and the successful removal of the original visual markers and text instructions.
    \item \textbf{Visual Realism:} Assesses the overall image quality, penalizing conspicuous artifacts, blurriness, or jagged edges to ensure natural blending.
    \item \textbf{Spatial Precision:} Evaluates whether the generated objects are complete and strictly confined within the spatial boundaries indicated by visual markers (e.g., bounding boxes) or follow specific directional arrows.
\end{itemize}

Based on these four scores, the evaluator outputs a structured JSON response containing a concise analysis and a final verdict of either \textbf{PASS} or \textbf{FAIL}. To achieve a PASS, a generated image must meet three strict conditions: an Instruction Fidelity score of $\ge 3.0$, an overall average score of $\ge 3.0$, and no single dimension scoring $\le 2.0$.

\begin{figure*}[htbp]
    \centering
    \includegraphics[width=\textwidth]{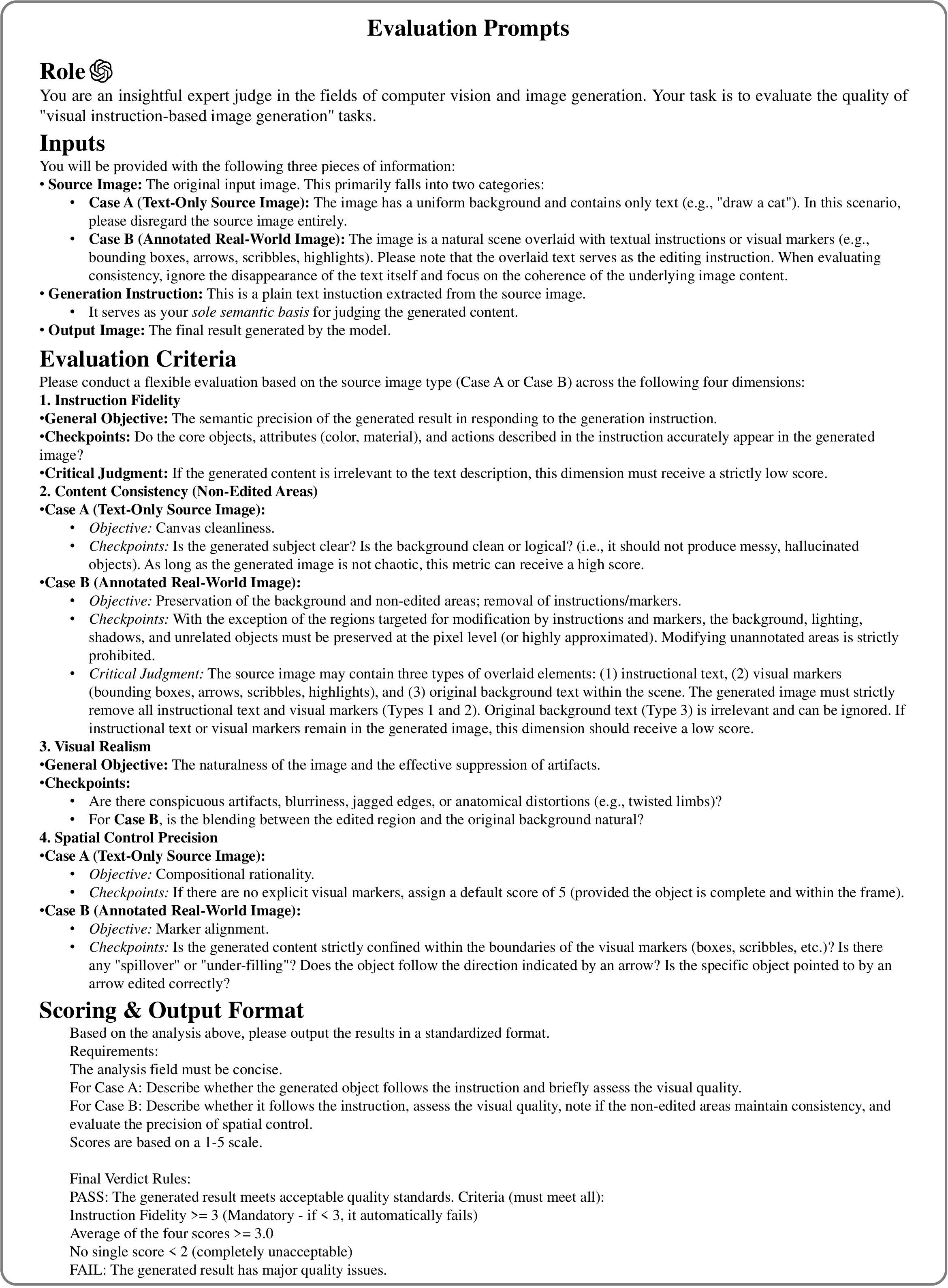}
    \caption{Evaluation prompts used in VLM evaluators.}
    \label{fig:eval_prompts}
\end{figure*}

For full transparency, the exact evaluation prompts detailing the role, criteria, and scoring rubrics used by the VLM evaluators are provided in Figure~\ref{fig:eval_prompts}. 
Furthermore, to guarantee a fair and standardized comparison across different models during the initial image generation phase, e evaluate each baseline using its optimal native interface, where the image inputs are supplemented with detailed text prompts expanded by Qwen3-VL.

\subsection{Quantitative and human evaluation}
\label{subsec:eval_details}

Building upon Section 3.2, we elaborate on the formulation of our tailored quantitative metrics and the human evaluation protocol.

\subsubsection{Quantitative Metrics Formulation}
Standard image generation metrics are inadequate for evaluating fine-grained, instruction-following visual edits. Therefore, we rigorously formulated the following similarity constraints:

\noindent\textbf{Directional CLIP Similarity:} 
To capture the semantic transition from the input to the generated image, we utilize the directional CLIP feature space. Let $c_{src}$ and $c_{tgt}$ be the MLLM-generated captions for the input image $I_{in}$ and the generated image $I_{gen}$, respectively. Using the CLIP text encoder $E_T$ and image encoder $E_I$, the metric is defined as:
\begin{equation}
    Sim_{\text{dir}} = \frac{(E_T(c_{tgt}) - E_T(c_{src})) \cdot (E_I(I_{gen}) - E_I(I_{in}))}{||E_T(c_{tgt}) - E_T(c_{src})|| \ ||E_I(I_{gen}) - E_I(I_{in})||}
    \label{eq:clip_dir}
\end{equation}
This measures how well the visual change aligns with the textual description of the edit.

\noindent\textbf{DINOv3 Directional Similarity ($\text{DINOv3 Sim}$):} 
While CLIP captures high-level semantics, DINOv3 is highly sensitive to dense spatial and physical structural changes. Let $\phi(\cdot)$ denote the dense feature extractor of DINOv3. We compute the cosine similarity of the edit displacement vectors between the generated output and the ground truth $I_{gt}$:
\begin{equation}
    Sim_{\text{DINOv3}} = \frac{(\phi(I_{gen}) - \phi(I_{in})) \cdot (\phi(I_{gt}) - \phi(I_{in}))}{||\phi(I_{gen}) - \phi(I_{in})|| \ ||\phi(I_{gt}) - \phi(I_{in})||}
    \label{eq:dinov3_sim}
\end{equation}
A higher $Sim_{\text{DINOv3}}$ strictly indicates that the model has accurately applied the spatial and physical transformations requested by the visual prompt, without introducing unintended background artifacts.

\subsubsection{Human Evaluation Protocol}
To complement the automated VLM-based assessments, we conducted a rigorous human evaluation involving 10 independent expert evaluators. 
We selected a stratified random subset of 250 diverse pairs from \benchname, ensuring a balanced representation of 25 samples per subset. 
During the evaluation, annotators were presented with the input visual instruction and the generated output. 
To avoid the inherent subjectivity of Likert scales, evaluators adopted a strict visual-inspection approach. 
Specifically, they first made a strict binary judgment on whether the generation overall qualified as a ``Pass'' or ``Fail''. 
If a sample was deemed a ``Fail'', the annotators were then required to explicitly record which specific dimensions out of the four criteria (Instruction Fidelity, Content Consistency, Visual Realism, Spatial Precision) contributed to the failure, allowing for multi-label tagging. To ensure a highly robust standard, all 250 samples were cross-checked by the independent evaluators to reach a reliable consensus.

\section{Model details}
\label{AppendG}
\subsection{Parameter breakdown}
\label{subsec:param_breakdown}
Table~\ref{tab:param_breakdown} reports the parameter breakdown of \modelname. 
The main trainable component is the Flow Backbone, which contains 1108.40M parameters and provides the core generative capacity. 
In addition, the Text-Image VAE contains 103.65M trainable parameters for mapping textual visual prompts into the image-like representation space. 
The Visual Encoder and Image VAE are kept frozen, with 322.58M and 83.65M parameters respectively, which reduces training cost while preserving strong visual representation and reconstruction priors.
\begin{table}[htbp]
\centering
\caption{\textbf{Parameter Breakdown of \modelname.}}
\label{tab:param_breakdown}
\small
\setlength{\tabcolsep}{4pt}
\renewcommand{\arraystretch}{1.0}
\begin{tabular}{lcc}
\toprule
\textbf{Component} & \textbf{Params (M)} & \textbf{Train.} \\
\midrule
Flow Backbone  & 1108.40 & Yes \\
Text-Image VAE & 103.65   & Yes \\
Visual Encoder & 322.58  & No  \\
Image VAE      & 83.65   & No  \\
\bottomrule
\end{tabular}
\end{table}

\subsection{Loss function for image in-image out generation}
\label{subsec:loss_func}
We jointly train our model for the image in-image out generation task using the following comprehensive optimization objective:

\begin{equation}
\label{eq:total_loss}
\mathcal{L} = \mathcal{L}_{\text{fm}} + \beta_1 \mathcal{L}_{\text{kld}} + \beta_2 \mathcal{L}_{\text{clip}}
\end{equation}

where $\beta_1$ and $\beta_2$ are the scaling weights for the KL-divergence loss and the contrastive loss, respectively. For the Flow Matching loss $\mathcal{L}_{\text{fm}}$, we compute the Mean Squared Error (MSE) between the predicted velocity field $v_\theta(z_t, t)$ at time-step $t$ and the ground-truth velocity $\hat{v}$.

To explicitly enforce semantic alignment within our visual-centric paradigm, we adapt a CLIP-style contrastive loss. Unlike traditional text-to-image models that align text and images, our model specifically aligns the unified visual instruction embedding with the generated image representation. Specifically, given a mini-batch of $N$ instruction-target pairs, we obtain the visual instruction latents $z_{TI}$ and extract the corresponding target image features $z_{I}$. We then compute the cosine similarity between all pairs of $z_{TI}$ and $z_{I}$ in the batch. This results in an $N \times N$ similarity matrix $S$, where each element $s_{i,j}$ represents the cosine similarity between the $i$-th instruction latent $z_{TI}$ and the $j$-th target image feature $z_{I}$. 

These similarity scores are subsequently scaled by a learnable temperature parameter $\tau$, denoted as $\text{logits}_{i,j} = s_{i,j} / \tau$. Following this, a symmetric cross-entropy loss over the similarity scores is computed along both the instruction-to-image and image-to-instruction directions:

\begin{equation}
\mathcal{L}_{\text{TI} \rightarrow \text{I}} = - \frac{1}{N} \sum_{i=1}^{N} \log \frac{\exp(\text{logits}_{i,i})}{\sum_{j=1}^{N} \exp(\text{logits}_{i,j})}
\end{equation}

\begin{equation}
\mathcal{L}_{\text{I} \rightarrow \text{TI}} = - \frac{1}{N} \sum_{i=1}^{N} \log \frac{\exp(\text{logits}_{i,i})}{\sum_{j=1}^{N} \exp(\text{logits}_{j,i})}
\end{equation}

We compute the average of these two components to obtain the final contrastive semantic alignment loss:

\begin{equation}
\mathcal{L}_{\text{clip}} = \text{CLIP}(z_{TI}, z_{I}) = \frac{1}{2} \left( \mathcal{L}_{\text{TI} \rightarrow \text{I}} + \mathcal{L}_{\text{I} \rightarrow \text{TI}} \right)
\end{equation}

For the KL divergence loss $\mathcal{L}_{\text{kld}}$, we regularize the visual instruction tokens towards a standard normal distribution $\mathcal{N}(0, 1)$ to prevent latent space collapse. Based on ablation study, we set the hyperparameters to $\beta_1 = 1 \times 10^{-2}$ and $\beta_2 = 1$.

\subsection{Hyperparameter ablation}
\label{hyperparms_abla}
\begin{table}[htbp]
\centering
\caption{Hyperparameter ablation study on \benchname. We report the average PASS Rate (\%) evaluated by three VLMs. Default optimal settings (CFG $= 7$, Steps $= 50$, $\beta_1 = 0.01$, $\beta_2 = 1$) are marked in \textbf{bold}. The control variable method is applied; parameters not actively ablated are fixed to their default values.}
\label{tab:hyper_ablation}
\setlength{\tabcolsep}{4pt} 
\begin{tabular}{@{}lc|lc|lc@{}}
\toprule
\multicolumn{2}{c|}{\textbf{(a) CFG Scale}} & \multicolumn{2}{c|}{\textbf{(b) Sampling Steps}} & \multicolumn{2}{c}{\textbf{(c) Loss Weights}} \\
\cmidrule(r){1-2} \cmidrule(lr){3-4} \cmidrule(l){5-6}
CFG & PASS (\%) & Steps & PASS (\%) & $(\mathcal{L}_{\text{clip}}, \mathcal{L}_{\text{kld}})$ & PASS (\%) \\
\midrule
1.1          & 38.5          & 10          & 25.7          & (0.1, 0.001)       & 46.7 \\
1.5          & 44.8          & 20          & 40.3          & (0.1, 0.01)        & 45.2 \\
3.0          & 47.7          & 30          & 46.3          & (0.1, 0.1)         & 44.2 \\
5.0          & 51.5          & 40          & 48.7          & (1, 0.001)         & 46.0 \\
\textbf{7.0} & \textbf{54.0} & \textbf{50} & \textbf{54.0} & \textbf{(1, 0.01)} & \textbf{54.0} \\
9.0          & 44.2          & -           & -             & (1, 0.1)           & 45.8 \\
\bottomrule
\end{tabular}
\end{table}
\noindent\textbf{Ablation on CFG scale.} The CFG scale dictates the degree to which the generated image aligns with the visual and textual instructions. We sweep the CFG scale from $1.1$ to $9$. 
As shown in Table~\ref{tab:hyper_ablation}, the PASS Rate exhibits an inverted U-shape trajectory. At lower scales (e.g., $1.1$ and $1.5$), the model struggles to strictly follow the editing constraints, yielding sub-optimal PASS rates ($38.5\%$ and $44.8\%$). 
The performance peaks at $54.0\%$ with a CFG scale of $7$. However, excessively high guidance (CFG $= 9$) degrades the generation quality, likely due to visual artifacts and color saturation typical of diffusion models, causing the PASS Rate to drop to $44.2\%$. Consequently, we adopt $7$ as the optimal CFG scale.

\noindent\textbf{Ablation in the sampling steps.} 
We evaluate the denoising process across various sampling steps ranging from $10$ to $50$. 
The results demonstrate a clear positive correlation between sampling steps and instruction adherence. 
Extreme low-step regimes (e.g., $10$ steps) yield a poor PASS Rate of $25.7\%$, indicating insufficient structural synthesis.
Performance improves substantially and begins to converge as steps increase, reaching our best result ($54.0\%$) in steps $50$. 
To balance computational efficiency and high-fidelity generation, $50$ sampling steps are selected for our primary evaluations.

\noindent\textbf{Ablation on loss weights.} 
The overall optimization objective is defined as $\mathcal{L} = \mathcal{L}_{\text{fm}} + \beta_1 \mathcal{L}_{\text{kld}} + \beta_2 \mathcal{L}_{\text{clip}}$, where $\mathcal{L}_{\text{fm}}$ (weight fixed to $1$) ensures structural integrity. 
We conduct a grid search over the KL divergence penalty $\beta_1$ and the CLIP semantic alignment weight $\beta_2$. 
The experiments reveal that the model is highly sensitive to this balance. 
Setting $\beta_2 = 1$ and $\beta_1 = 0.01$ achieves the superior PASS Rate of $54.0\%$. Lowering the CLIP weight (e.g., $\beta_2 = 0.1$) universally harms performance (hovering around $45\% - 46\%$), as it weakens the semantic alignment between the visual instruction embedding $z_{TI}$ and the image embedding $z_I$.
Conversely, over-penalizing the KL divergence ($\beta_1 = 0.1$) overly restricts the latent space, hindering the model's expressive capacity.

\subsection{Experimental details}
\label{exp_details}
We utilize the WebDataset~\cite{aizman2020highperformanceiolarge} format for highly efficient, streaming-based I/O processing of large-scale image-instruction pairs. 
To prevent the model from overfitting to dominant task categories and to ensure stable gradient descent, we employ a strictly balanced mini-batch sampling strategy. 
Specifically, this strategy guarantees that the eight distinct dataset sub-categories are uniformly distributed within each training batch, effectively mitigating task-level bias and ensuring balanced optimization across all instruction types.
Initialized with pre-trained weights from Crossflow~\cite{liu2025flowing}, our 1.2B-parameter model, \modelname, benefits from strong prior knowledge of visual semantics. 
During training, the resolution of the output images is $256 \times 256$. 
We employ the AdamW optimizer with a base learning rate of $1 \times 10^{-4}$ alongside a standard cosine decay schedule with a linear warmup.
The entire training process was conducted for 240,000 steps with a global batch size of 512, requiring approximately 240 A100 GPU hours to reach convergence. 
During the evaluation phase, when generating images on \benchname for comparison, all baseline models are executed using their official default inference parameters to ensure a standardized and fair assessment.

\end{document}